\pdfoutput=1

\documentclass[11pt]{article}

\usepackage[table]{xcolor}

\usepackage{acl}

\usepackage{graphicx}
\usepackage{times}
\usepackage{booktabs}
\usepackage{amsmath}
\usepackage{nicematrix}
\usepackage[nameinlink,capitalize,noabbrev]{cleveref}
\usepackage{amssymb}
\usepackage{latexsym}
\usepackage{multicol}
\usepackage{subcaption}
\usepackage{multirow}
\usepackage{bm}
\usepackage{todonotes}
\usepackage[frozencache,cachedir=.]{minted}
\usepackage[T1]{fontenc}

\usepackage[utf8]{inputenc}

\usepackage{microtype}
\usepackage{csquotes}

\usepackage{tikz}
\newcommand*\circled[1]{\tikz[baseline=(char.base)]{
            \node[shape=circle,draw,inner sep=2pt] (char) {#1};}}

\title{\textit{More Room for Language:} \\
Investigating the Effect of Retrieval on Language Models}

\author{David Samuel \And Lucas Georges Gabriel Charpentier\\ \\Language Technology Group, University of Oslo \And Sondre Wold \\
}

\begin{document}
\maketitle
\begin{abstract}
Retrieval-augmented language models pose a promising alternative to standard language modeling. During pretraining, these models search in a corpus of documents for contextually relevant information that could aid the language modeling objective. We introduce an \textit{`ideal retrieval'} methodology to study these models in a fully controllable setting. We conduct an extensive evaluation to examine how retrieval augmentation affects the behavior of the underlying language model. Among other things, we observe that these models: \emph{i)}~save substantially less world knowledge in their weights, \emph{ii)}~are better at understanding local context and inter-word dependencies, but \emph{iii)}~are worse at comprehending global context.
\end{abstract}

\section{Introduction}

Retrieval-augmented language models combine the strengths of self-supervised pretraining with information retrieval techniques to allow for information extraction from a nonparametric memory. During pretraining, the prediction of masked tokens is conditioned not only on the immediate context but also on information found to be contextually relevant by a similarity search over a knowledge database. These models are typically proven effective in knowledge-intensive tasks, such as answering open-domain questions \citep{pmlr-v119-guu20a, lewis-etal-2022-boosted, izacard2023atlas}. 

However, little emphasis has been put into understanding what this type of training scheme does to the underlying language model when analyzed as a standalone -- separated from the overall retrieval pipeline. 
Retrieval augmentation is often proposed as a better alternative to standard pretraining, without much evidence of its advantages and disadvantages. 
The behavior of the entire pipeline depends on the qualities of the retrieved database and the qualities of the standalone language model. While the database is relatively easy to control, the performance of the language model can be hard to estimate. This paper aims to shed more light on the expected qualities of the language model, separated from the database retrieval.

In total, we evaluate the effect of retrieval on 9 language models with 8 sets of zero-shot, probing and finetuning tasks to empirically show that:
\begin{enumerate}
    \item \textbf{Retrieval augmentation separates linguistic knowledge from world knowledge}, to some extent -- the language model alone improves syntactic understanding while delegating world knowledge to the retrieval module. This separation becomes larger with scale.
    \item \textbf{Retrieval augmentation negatively impacts NLU performance} -- the stand-alone language model performs worse in multi-sentence language understanding, which is concerning for general-use language models.
    \item \textbf{Poor retrieval quality does not negatively impact pretraining} -- the model behavior gets closer to the baseline no-retrieval performance, without overall quality degradation.
\end{enumerate}

\begin{figure}[t!]
    \centering
    \includegraphics[width=\columnwidth]{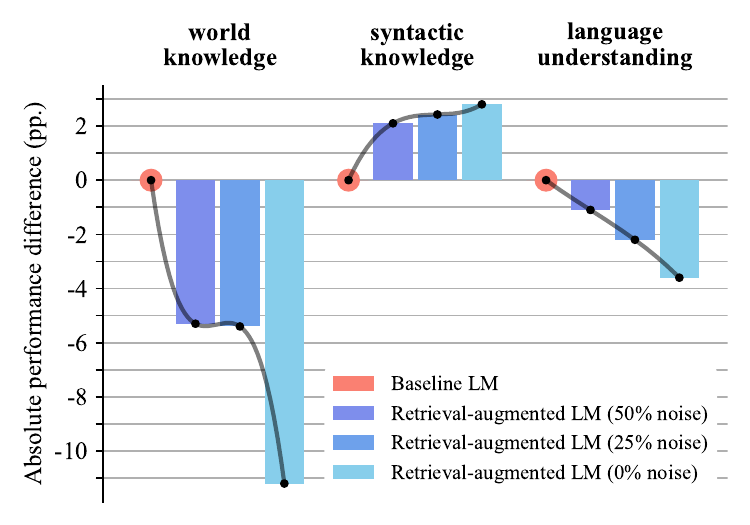}
    \caption{The aggregated absolute differences from the baseline across three categories of benchmarks, the models exhibit consistent differences for each category.}
    \label{fig:aggregated-performance}
\end{figure}

\section{Related work}

\paragraph{Evaluation of retrieval augmentation}
While there has been a lot of effort put into developing different retrieval-augmented language models \citep{pmlr-v119-guu20a, borgeaud2022improving, izacard2023atlas}, little emphasis has been put on analyzing the limitations and abilities of current methods. Recently, \citet{norlund-etal-2023-generalization} found that the reliance on surface-level similarities between the retrieval database and test data has been somewhat understated in the literature, finding that token-level overlap accounts for some of the reported performance in the popular \textsc{RETRO} architecture \citep{borgeaud2022improving}; similarly, \newcite{wang-etal-2023-knn} shows that even though retrieval augmentation improves the perplexity of language models, it does not improve their generation quality. Some have focused on the retrieval part of the pipeline, with \citet{doostmohammadi2023surfacebased} reporting that a sparse retrieval index can decrease perplexity for retrieval-augmented language models. 
\citet{charpentier-etal-2023-brent} found that retrieval-augmented pretraining can improve context utilization.

\paragraph{From-scratch pretraining}

Most current retrieval-augmented models are created by finetuning or continual training (retrofitting) of an already pretrained model. As shown in \newcite{wang-etal-2023-shall}, only RETRO trains a retrieval-augmented model from scratch.
While \citet{borgeaud2022improving} focus on improving the perplexity and text generation with retrieval assistance, we want to look at whether pretraining with retrieval leads to models having better syntactic understanding while retaining less world knowledge. This builds on the intuition that retrieval should free up parameter space for linguistic knowledge, as the relevant world-knowledge information is continuously supplied in the retrieved input. This hypothesis can be tested only by pretraining a blank model from scratch.

\section{Controlled retrieval augmentation}
\label{sec:method}

\begin{figure*}[!t]
    \centering
    \includegraphics[width=1.0\textwidth]{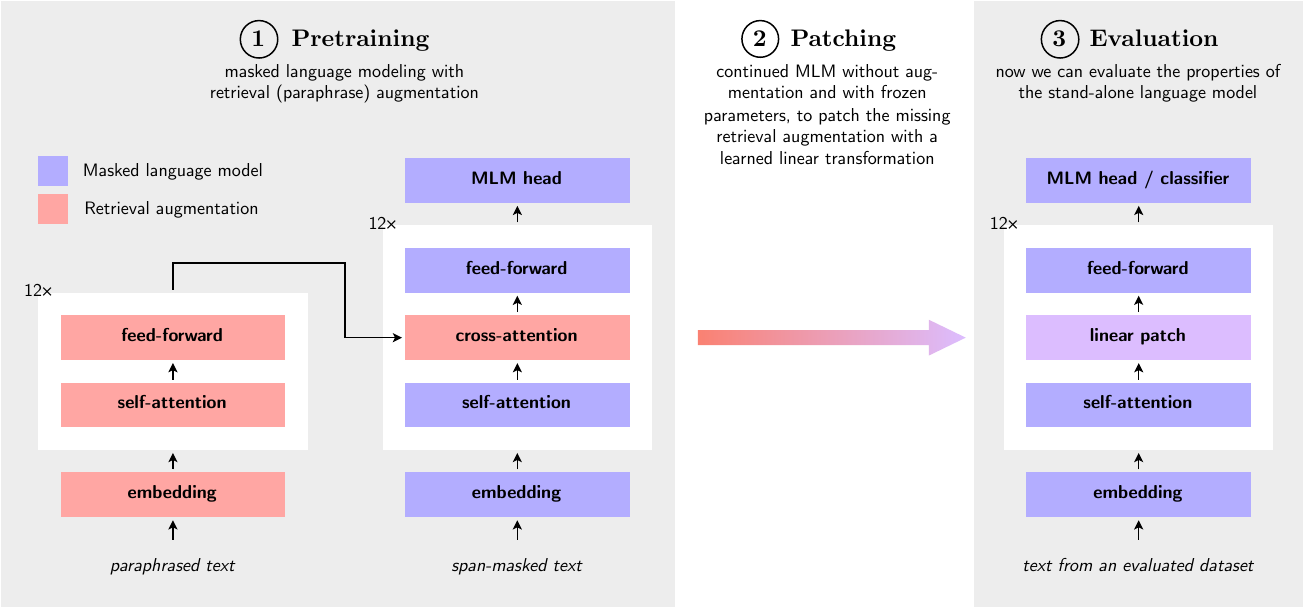}
    \caption{Illustration of the full evaluation process. {\small\circled{1}} Pretraining of a retrieval-augmented language model, using an encoder-decoder transformer architecture. The retrieval mechanism is greatly simplified with paraphrase-based retrieval augmentation. The language model learns to recover a partially masked text while having access to its unmasked but paraphrased version. {\small\circled{2}} To evaluate the standalone language model (in blue), we have to remove the retrieval augmentation (in red) and replace the cross-attention module with a simple linear projection (in purple). {\small\circled{3}} Only the patched language model is used during the evaluation to investigate its stand-alone features.}
    \label{fig:rer-diagram}
\end{figure*}

This study examines the general implications of retrieval augmentation in language modeling, in a fully controllable `laboratory' setting and without relying on a particular retrieval model or parameters. All existing retrieval models are noisy (not always retrieving relevant context) and the noise might not only have a large impact on the downstream performance but also it is hard to measure and control. Therefore, we use an impractical,\footnote{As in `only useful for a theoretic study'.} but fully controllable, \textit{perfect retrieval} in the form of paraphrased inputs, as illustrated in \cref{fig:rer-diagram}. Our goal is to study the effect of retrieval augmentation on the stand-alone language model, and this setup allows us to separate the effect of retrieval type, retrieval accuracy or frequency of retrieved duplicates. As discussed later, we also provide the results of retrieval augmentation with a controlled amount of noise to get closer to a realistic scenario.

\paragraph{Simplified retrieval-augmented LM} We base our experiments on \textit{masked language models} as they offer greater flexibility for evaluation \citep{devlin-etal-2019-bert, rogers-etal-2020-primer}. The retrieval augmentation is substantially simplified thanks to paraphrase-based pretraining. As a whole, the model is an encoder-decoder transformer \citep{NIPS2017_3f5ee243}, where the encoder embeds the retrieved context and the decoder is a language model (\cref{fig:rer-diagram}). Specifically, the decoder is given masked text segments, its training objective is to unmask it \citep{devlin-etal-2019-bert} and the encoder is provided with a paraphrase of the unmasked segment.

\paragraph{Paraphrased training data} We utilize the English Wikipedia as a clean and information-rich text corpus. Due to the cost of paraphrasing, we select only the top 10\% most visited articles by page view count in the last year (about 400 million words). The paraphrases are generated by a prompted instruction-tuned Mistral 7B language model \citep{jiang2023mistral}, as described in \cref{app:mistral}.\footnote{Such a dataset might be useful for tasks outside the scope of this paper and we openly release it at \url{https://huggingface.co/datasets/ltg/en-wiki-paraphrased}.}

\paragraph{Quality of paraphrases}
\label{sec:paraphrase-quality}

It is essential to train the models on \textit{good} paraphrases to avoid retrieval of irrelevant context and unwanted data leakage. For this study, a \textit{good} paraphrase should have the same meaning as the original but be completely different lexically and syntactically. In that way, the retrieval can provide relevant context and world knowledge without inhibiting the training signal by allowing the model to simply copy the paraphrased document word-by-word.

Firstly, we utilize deep contextualized sentence embeddings to measure the preservation of meaning as the average semantic similarity of every original-paraphrase pair \citep{reimers-gurevych-2019-sentence}. Specifically, the average cosine similarity is 0.88,\footnote{According to \texttt{all-mpnet-base-v2}, the best \texttt{SentenceTransformers} model as of December 2023: \url{https://www.sbert.net/docs/pretrained_models.html}.} indicating that the paraphrases are almost semantically identical to the original texts.\footnote{As a reference, note that the illustrative example in \cref{fig:rer-diagram} has a slightly lower semantic similarity of 0.85.}

Secondly, we measure the lexical (and to some extent syntactic) similarity as the BLEU score between paraphrased and original texts \citep{papineni-etal-2002-bleu, post-2018-call, niu-etal-2021-unsupervised}. The average BLEU score is 13\% for the raw pairs and 7\% for pairs with removed named entities and digits -- this shows that the paraphrases should not leak surface-level information.

\paragraph{Noise injection} The paraphrasing allows us to test the effect of a perfectly accurate retriever. However, a real retriever does not always provide relevant context. To also evaluate a more realistic retrieval setting, we sometimes inject a randomly sampled context, according to a given noise probability.

\paragraph{Linear patching} We need to separate the language model from its retrieval augmentation to measure its independent performance. However, when removed naively, the separated language model exhibits poor performance because it expects nonzero vectors from the cross-attention mechanism. Therefore, we replace the retrieval augmentation with a simple linear layer and continue pretraining with all other parameters frozen, as illustrated in \cref{fig:rer-diagram}. In \cref{app:patch}, we empirically prove that \textit{(i)} the patching is necessary and that \textit{(ii)} the linear patches are weak enough to not provide additional knowledge.

\renewcommand{\arraystretch}{1.3}

\begin{table*}[!ht]

\resizebox{\textwidth}{!}{%
    \begin{NiceTabular}{@{}l !{\qquad} w{c}{4em} w{c}{4em} w{c}{4em} w{c}{0.5em} w{c}{4em} w{c}{4em} w{c}{4em} w{c}{4em} w{c}{0.5em} w{c}{4em} W{c}{4em} w{c}{4em}@{}}
\CodeBefore
\cellcolor[HTML]{fffefe}{8-2}
\cellcolor[HTML]{fffefe}{8-3}
\cellcolor[HTML]{fffefe}{8-4}
\cellcolor[HTML]{dadaff}{8-6}
\cellcolor[HTML]{ceceff}{8-7}
\cellcolor[HTML]{f0f0ff}{8-8}
\cellcolor[HTML]{c6c6ff}{8-9}
\cellcolor[HTML]{fffefe}{8-11}
\cellcolor[HTML]{fffefe}{8-12}
\cellcolor[HTML]{fffefe}{8-13}
\cellcolor[HTML]{ffeaea}{9-2}
\cellcolor[HTML]{ffd8d8}{9-3}
\cellcolor[HTML]{ffe4e4}{9-4}
\cellcolor[HTML]{eeeeff}{9-6}
\cellcolor[HTML]{ececff}{9-7}
\cellcolor[HTML]{dcdcff}{9-8}
\cellcolor[HTML]{fffefe}{9-9}
\cellcolor[HTML]{ffeaea}{9-11}
\cellcolor[HTML]{fffafa}{9-12}
\cellcolor[HTML]{fff6f6}{9-13}
\cellcolor[HTML]{ffecec}{10-2}
\cellcolor[HTML]{ffdada}{10-3}
\cellcolor[HTML]{ffe2e2}{10-4}
\cellcolor[HTML]{f0f0ff}{10-6}
\cellcolor[HTML]{f3f3ff}{10-7}
\cellcolor[HTML]{ececff}{10-8}
\cellcolor[HTML]{ececff}{10-9}
\cellcolor[HTML]{ffe2e2}{10-11}
\cellcolor[HTML]{fff8f8}{10-12}
\cellcolor[HTML]{fff2f2}{10-13}
\cellcolor[HTML]{ffe2e2}{11-2}
\cellcolor[HTML]{ffcccc}{11-3}
\cellcolor[HTML]{ffd2d2}{11-4}
\cellcolor[HTML]{fffefe}{11-6}
\cellcolor[HTML]{fffefe}{11-7}
\cellcolor[HTML]{fffefe}{11-8}
\cellcolor[HTML]{fffefe}{11-9}
\cellcolor[HTML]{ffdada}{11-11}
\cellcolor[HTML]{fff2f2}{11-12}
\cellcolor[HTML]{fff0f0}{11-13}
\cellcolor[HTML]{fffefe}{13-2}
\cellcolor[HTML]{fffefe}{13-3}
\cellcolor[HTML]{fffefe}{13-4}
\cellcolor[HTML]{f6f6ff}{13-6}
\cellcolor[HTML]{eaeaff}{13-7}
\cellcolor[HTML]{f3f3ff}{13-8}
\cellcolor[HTML]{e0e0ff}{13-9}
\cellcolor[HTML]{fffefe}{13-11}
\cellcolor[HTML]{eeeeff}{13-12}
\cellcolor[HTML]{fffefe}{13-13}
\cellcolor[HTML]{ffe2e2}{14-2}
\cellcolor[HTML]{ffd0d0}{14-3}
\cellcolor[HTML]{ffd3d3}{14-4}
\cellcolor[HTML]{fffefe}{14-6}
\cellcolor[HTML]{fffefe}{14-7}
\cellcolor[HTML]{fffefe}{14-8}
\cellcolor[HTML]{fffefe}{14-9}
\cellcolor[HTML]{ffd8d8}{14-11}
\cellcolor[HTML]{fffefe}{14-12}
\cellcolor[HTML]{ffecec}{14-13}
\cellcolor[HTML]{fffefe}{16-2}
\cellcolor[HTML]{fffefe}{16-3}
\cellcolor[HTML]{fffefe}{16-4}
\cellcolor[HTML]{f0f0ff}{16-6}
\cellcolor[HTML]{ceceff}{16-7}
\cellcolor[HTML]{fffefe}{16-8}
\cellcolor[HTML]{fffefe}{16-9}
\cellcolor[HTML]{fffefe}{16-11}
\cellcolor[HTML]{e8e8ff}{16-12}
\cellcolor[HTML]{fffefe}{16-13}
\cellcolor[HTML]{ffecec}{17-2}
\cellcolor[HTML]{ffe6e6}{17-3}
\cellcolor[HTML]{ffcccc}{17-4}
\cellcolor[HTML]{fffefe}{17-6}
\cellcolor[HTML]{fffefe}{17-7}
\cellcolor[HTML]{fffafa}{17-8}
\cellcolor[HTML]{fff2f2}{17-9}
\cellcolor[HTML]{ffe0e0}{17-11}
\cellcolor[HTML]{fffefe}{17-12}
\cellcolor[HTML]{ffecec}{17-13}
\Body
    \toprule
    \Block{4-1}{\textbf{Model}} & \Block{1-3}{\textsc{world knowledge}} & & & & \Block{1-4}{\textsc{syntactic knowledge}} & & & & & \Block{1-3}{\textsc{language understanding}} \\
    & \Block{2-1}{\textbf{Concept}\\\textbf{Net}}&  \Block{2-1}{\textbf{SQuAD}}&  \Block{2-1}{\textbf{TREx}} & & \Block{2-1}{\textbf{linear}\\\textbf{probing}} &  \Block{2-1}{\textbf{attention}\\\textbf{probing}}&  \Block{2-1}{\textbf{BLiMP}} &  \Block{2-1}{\textbf{MSGS}} & &  \Block[c]{2-1}{\textbf{\kern-0.4emLAMBADA}} &  \Block{2-1}{\textbf{GLUE}} &  \Block{2-1}{\textbf{SQuAD}} \\
    &  \\
         & \footnotesize{(MRR $\uparrow$)} & \footnotesize{(MRR  $\uparrow$)} & \footnotesize{(MRR $\uparrow$)}  & & \footnotesize{(LAS $\uparrow$)} & \footnotesize{(UUAS  $\uparrow$)} & \footnotesize{(Acc. $\uparrow$)} & \footnotesize{(LBS $\uparrow$)} & & \footnotesize{(Acc. $\uparrow$)} & \footnotesize{(Avg.  $\uparrow$)} & \footnotesize{(F\textsubscript1 $\uparrow$)} \\

    \midrule
    
    \textsc{reference model (110m)}& & & \\
 \hspace{1em}\textit{bert-base-cased}& \textit{26.0}& \textit{34.0}&\textit{62.0} & & \textit{82.0} & \textit{45.1} & \textit{85.6} & \textit{-0.10} & & \textit{44.8} & \textit{82.1} & \textit{88.4} \\[0.5em]
 \textsc{base (98m)}& & &\\
 \hspace{1em}$\bm{-}$ retrieval & \textbf{20.3}& \textbf{32.1}&\textbf{53.6} & & 78.1 & 48.0  & 82.9 & -0.47 & & \textbf{46.0} & \textbf{82.2} & \textbf{91.2} \\
 \hspace{1em}$\bm{+}$ retrieval (50\% noise) & 17.7 & 23.2 & 49.1 & & 79.8 & 51.3 & 81.3 & \textbf{-0.37} & & 43.2 & 82.0 & 90.7\\
 \hspace{1em}$\bm{+}$ retrieval (25\% noise) & 18.1 & 23.4 & 48.3 & & 79.9 & 51.6 & 82.7 & -0.38 & & 40.6 & 81.9 & 90.2\\
 \hspace{1em}$\bm{+}$ retrieval (0\% noise) & 14.9 & 15.8 &41.5 & & \textbf{80.2} & \textbf{51.8} & \textbf{83.2} & \textbf{-0.37} & & 37.5 & 81.2 & 89.7\\[0.5em]
 \textsc{small (28m)}& & &\\
\hspace{1em}$\bm{-}$ retrieval & \textbf{17.2} & \textbf{28.3} & \textbf{47.4} & &  71.2 & 49.7 & 78.6 & -0.56 & & \textbf{35.1} & 78.0 & \textbf{88.6}\\
\hspace{1em}$\bm{+}$ retrieval & 11.8 & 15.3 & 36.3 & &\textbf{71.7} & \textbf{50.4} & \textbf{78.8} & \textbf{-0.53} & & 26.2 & \textbf{78.4} & 86.2\\[0.5em]
 \textsc{x-small (9m)}& & &\\
\hspace{1em}$\bm{-}$ retrieval & \textbf{9.9}& \textbf{14.7}&\textbf{39.2} & & 63.3 & 45.5 & \textbf{73.4}  & \textbf{-0.55} & & \textbf{25.3} & 75.2 & \textbf{81.1}\\
\hspace{1em}$\bm{+}$ retrieval & 7.5 & 10.6& 23.4 & &\textbf{63.6} & \textbf{49.2} & 73.3 & -0.57 & & 19.3 & \textbf{76.0} & 78.7\\
   \bottomrule 
    \end{NiceTabular}%
     }
     \caption{The overall evaluation scores for all sets of tasks, are divided into three categories. $\bm{+}$ denotes models pretrained with retrieval augmentation while $\bm{-}$ denotes standard models pretrained without retriever; note that the evaluation is done without any retrieval mechanism for all models (see \cref{sec:eval-method}). We divide the models into three subsets based on their size and also give the reference scores of the official \texttt{bert-base-cased} model evaluated with our pipeline. We highlight the best results for each model size in \textbf{boldface} and measure the average score across 5 runs, when applicable. The red color indicates worse results than the no-retrieval baseline and vice-versa for the blue color.}
     \label{tab:results}
\end{table*}

\section{Evaluation}

The experiments in this section evaluate how retrieval augmentation, size and retrieval quality affect world knowledge, syntactic knowledge and language understanding of language models.

\paragraph{Evaluated language models} 

We follow the LTG-BERT architecture and training choices for pretraining the masked language models; this method is designed to work competitively in low-resource settings, making it suitable for our study \citep{samuel-etal-2023-trained}. In total, we pretrain eight models: three sizes: \textsc{x-small} (8.5M parameters), \textsc{small} (27.7M) and \textsc{base} (98.2M), and each size with \& without retrieval augmentation. We also experiment with the noise injection for the \textsc{base} model, we train two additional models with 25 and 50\% noise probability. The pretraining details are listed in \cref{app:pretraining}. We openly release all pretrained models, as well as the training code, online.\footnote{\url{https://github.com/ltgoslo/more-room-for-language}}

\paragraph{Evaluation method} \label{sec:eval-method} As stated previously, our objective is to evaluate the effect of pretraining with retrieval augmentation on a standalone language model. Therefore, all evaluation tests are performed without any retrieval mechanism and on tasks that do not benefit from retrieval. We use linear patching (\cref{fig:rer-diagram}) to remove the retrieval augmentation.

\paragraph{World knowledge}

To evaluate the knowledge capacity of a language model, we evaluate it in a zero-shot setting on the Language Model Analysis probe \citep[LAMA;][]{petroni-etal-2019-language}. The probe provides cloze-style statements of factual information from different sources. We evaluate all models on the subsets extracted from SQuAD \citep{rajpurkar-etal-2016-squad}, from the ConceptNet knowledge graph \citep{speer2017conceptnet} and from the Wikipedia-based T-REx \citep{elsahar2018t}.

\paragraph{Syntactic knowledge}

There are many ways of measuring the syntactic understanding of a language model, each with its disadvantages \citep{belinkov-2022-probing}. We aim for a robust evaluation and thus measure the syntactic knowledge on four different types of benchmarks. First, with \textit{linear probing}, we test how easy it is to extract syntactic dependencies between words from contextualized subword embeddings \citep{shi-etal-2016-string, DBLP:conf/iclr/AlainB17, liu-etal-2019-linguistic}. Second, \textit{attention probing} measures how well we can construct dependency trees directly from attention probabilities \citep{marecek-rosa-2018-extracting, raganato-tiedemann-2018-analysis, ravishankar-etal-2021-attention}. Then \textit{BLiMP} tests if a language model prefers well-formed grammatical sentences \citep{warstadt-etal-2020-blimp-benchmark, salazar-etal-2020-masked}. Finally, \textit{MSGS} leverages the poverty of the stimulus
design \citep{wilson-2006} to measure the level of linguistic generalization \citep{warstadt-etal-2020-learning}.

\paragraph{Language understanding} The third set of benchmarks evaluates different aspects of general language understanding.   \textit{LAMBADA} tests the ability to understand long passages of text and form long-range dependencies \citep{paperno-etal-2016-lambada}. \textit{GLUE} is a multitask benchmark for fine-tuning and evaluating language models on diverse downstream tasks \citep{wang-etal-2018-glue}. \textit{SQuAD} measures the degree of reading comprehension using an extractive question-answering dataset \citep{rajpurkar-etal-2016-squad}.

\paragraph{Results} We present the overall results in \cref{tab:results} and \cref{fig:aggregated-performance}. Fine-grained per-task results with significance tests (when applicable), and an in-depth explanation of the evaluated tasks and our setup are provided in \cref{app:evaluation}.

\section{Discussion}

\paragraph{Retrieval augmentation separates linguistic knowledge from world knowledge} There is a clear trend in the evaluated performance between the world knowledge tasks and the linguistic tasks -- when a language model can rely more on retrieval during pretraining (with decreased retrieval noise), it remembers fewer facts and gets progressively worse on all evaluated world knowledge tasks; but, on the other hand, its syntactic understanding consistently improves (\cref{tab:results}).

This strongly suggests that a language model with retrieval does not allocate as many parameters to store world knowledge and instead uses the freed parameters for other features, such as understanding syntax. As a result, retrieval-augmented pretraining leads to a clear separation between world knowledge (in the retriever) and syntactic knowledge (in the language model). This modular system allows one to easily update factual knowledge by updating the retrieval database, without risking any side effects from updating neural parameters \citep{de-cao-etal-2021-editing, yao2023editing}.

The positive results on syntactic tasks suggest that retrieval-based pretraining can be a promising avenue for efficient language modeling, as far as the goal is to train a small model that understands syntax well. Another notable finding is that the linguistic advantages of retrieval-pretrained models over standard models grow with the size of these models (\cref{tab:results}).

\paragraph{Retrieval augmentation negatively impacts NLU performance} Contrary to the mostly local syntactic understanding, language understanding gets worse with retrieval-augmented pretraining (\cref{tab:results}). The fine-grained GLUE results in \cref{tab:glue} show that this affects tasks that require global inter-sentence comprehension tasks (NLI) more than short-range local tasks (CoLA or SST-2).

We argue that this is in part caused by the lack of factual knowledge (which can help to resolve ambiguous cases), but that it is also indirectly caused by the way retrieval-augmented pretraining works. When searching for the global context, the language model is incentivized to trust the retrieved document more than the partially masked input, which pushes the mechanism of long-range resolution out of the language model itself. We further investigate this in an additional experiment in \cref{app:long_context}, where we also utilize the retriever augmentation during evaluation -- this setting clearly improves the performance of long-range context resolution on LAMBADA, which suggests that the processing of global context is mainly delegated out of the language model itself to its retrieval augmentation.

This behavior poses a challenge to using retrieval augmentation for pretraining general-purpose language models. It makes retrieval finetuning (as opposed to full pretraining) not only less costly but also a more performant alternative.

\paragraph{Poor retrieval quality does not negatively impact pretraining} Noisy retrieval pretraining does not lead to an overall drop in performance; instead, it interpolates the behavior of standard pretraining and of pretraining with a perfect retrieval (\cref{tab:results}) -- more noise makes the retrieved context less reliable and the language model has to act more independently, akin to the standard no-retrieval setting.

While a high-quality retrieval mechanism is critical during inference, our results could suggest that a subpar (but computationally inexpensive) retrieval during training does not negatively impact the overall performance.

\section{Conclusion}

We introduced a novel theoretical framework for studying the properties of retrieval-augmented language models. Specifically, through this paper, we were able to show that using retrieval during pretraining leads models to learn less world knowledge while gaining better syntactic knowledge; this separation is especially pronounced for larger models. However, this improvement comes at the cost of performance in general language understanding and in resolving long-range context dependency. Due to the model relying on the retrieved spans, the global context resolution seems to be delegated to the retrieval module. We also performed an ablation on the effect of noisy retrieval and saw that it only slightly affects the syntactic capabilities of the model while substantially improving both its language understanding skills and world knowledge. We make all resources used in the paper openly available at {\footnotesize\url{https://github.com/ltgoslo/more-room-for-language}}.

\clearpage

\section*{Limitations}

\paragraph{Pretraining corpus} We pretrain all language models on the texts from the English Wikipedia -- which is an information-rich and high-quality corpus, but also one that is monolingual and not very stylistically diverse. More typical web-crawl-based corpora are not as rich in factual information and the differences in evaluation of world knowledge might not be as pronounced for them. Similarly, we only evaluate the syntactic knowledge of an English knowledge model, and the results might differ for a typologically different language.

\paragraph{Model scale} Due to our computational constraints, we decided to limit the size of the pretrained language models to 100M parameters. While our results show a consistent trend from the smallest to the largest models, there is a possibility that this suddenly changes in the billion-parameter scale.

\paragraph{Masked language modeling} This study only evaluates the performance of masked language models, primarily due to a larger and more diverse set of benchmarks and also due to their ability to perform well even at a modest scale. We believe that most of our findings hold for causal language models, too; and we look forward to future work that evaluates these (typically much larger) models.

\paragraph{Need of paraphrasing} Our evaluation framework relies on pretraining on a parallel corpus of masked texts and their paraphrases. Generating the paraphrases by autoregressive decoding from a language model is costly; in turn, this cost limits the size of the pretraining corpus that can be tested. This is why we decided to use a relatively small corpus of 400M words in this work.

\section*{Acknowledgements}

We would like to thank Erik Velldal, Andrey
Kutuzov and Lilja Øvrelid for providing valuable
feedback on this work and for their never-ending support. The final version is written more clearly thanks to the feedback from Egil Rønningstad and Yves Scherrer.

The computations were performed on resources
provided through Sigma2 – the national research
infrastructure provider for High-Performance Com-
puting and large-scale data storage in Norway.

\bibliography{anthology,custom}
\bibliographystyle{acl_natbib}

\clearpage
\onecolumn
\appendix

\setlength{\parindent}{0pt}
\setlength{\parskip}{0.7em}

\section{Mistral 7B paraphrase generation}
\label{app:mistral}

We use the \texttt{mistralai/Mistral-7B-Instruct-v0.1} checkpoint available on HuggingFace \citep{wolf-etal-2020-transformers}.\footnote{Online link: \url{https://huggingface.co/mistralai/Mistral-7B-Instruct-v0.1}} We use a combination of nucleus sampling and top-$k$ sampling with temperature of $0.9$, $p$ of $0.9$ and $k$ of $50$ \citep{Holtzman2020The}. We also utilize a repetition penalty of $1.2$ to avoid degenerate outputs. We release the resulting parallel corpus online.\footnote{Available on HuggingFace: \url{https://huggingface.co/datasets/ltg/en-wiki-paraphrased}}

The instructive prompt template for paraphrasing is the following Python f-string:
\begin{minted}[linenos=true, breaklines]{python}
"""<s>[INST] Paraphrase the following paragraphs, try to be very creative and make it look as different as possible without changing any meaning or losing any information. Don't be afraid to change the order of words or sentences. Don't add any new information that is not already in the text.

{original_unmasked_segment} [/INST]"""
\end{minted}

\paragraph{Digression: issues of safety alignment} We chose Mistral because of its state-of-the-art size-to-power performance \citep{jiang2023mistral}, and its consistent and reliable outputs -- but also because of its light safety alignment. Other, much more safety-aligned language models, sometimes refused to paraphrase Wikipedia articles about problematic topics, such as the development of nuclear bombs or the biography of Adolf Hitler; yet we have not encountered such issues with Mistral. We note that while there are good arguments for restricting the capabilities of language models for the general public, the `safe' language models can be limiting for scientific research. As an example, this is the output of Llama-2-7b by \newcite{touvron2023llama} when asked to paraphrase a seemingly controversial Wikipedia article:
\renewcommand\mkblockquote[4]{\enquote{#1#2#3}#4}
\blockquote{\textit{It is not appropriate or ethical to provide information or assistance that promotes or glorifies harmful or hateful ideologies, individuals, or actions. Therefore, I cannot paraphrase the given text about Adolf Hitler as it would be inappropriate and potentially harmful to do so. Instead, I suggest focusing on topics that promote positivity, inclusivity, and respect for all individuals. Is there anything else I can assist you with?}}


\section{Effect of linear patching}
\label{app:patch}

As discussed in \cref{sec:method}, we have to apply a linear patch in order to conduct a fair evaluation of the separated language model, the whole process is also illustrated in the following figure -- we add a liner layer (called a linear patch) between the self-attention and feed-forward network of each layer of the encoder as a proxy to the missing cross-attention:

\vspace{1em}

The purpose of this section is to empirically show that the linear patching is necessary and that it does not bias the results by providing any additional knowledge to the patched model. For that, we will use the detailed results from \cref{app:evaluation} that compare the performance of the patched and unpatched models.

\subsection{Patching is necessary for the retrieval models}

The results clearly show that when we evaluate the separated language model pretrained with retrieval, it completely fails without patching when evaluated on tasks that do not involve any finetuning. While this effect is clear across all tasks (\cref{app:lama,app:lp,app:blimp}), we will illustrate it specifically on the LAMBADA task from \cref{app:lambada}. There, the \textsc{x-small}, \textsc{small} and \textsc{base} retrieval models achieve 0\%, 0\% and 23\% accuracy without a patch, which is substantially less than the 19\%, 26\% and 38\% accuracy with a simple linear patch. The naive removal of the cross-entropy modules (\cref{fig:rer-diagram}) hinders the language model and the linear patching is able to remove this handicap. Note that the naive removal is not a problem for a model that is further finetuned -- for example, the no-patch $to$ patch SQuAD F\textsubscript1 scores stay very stable for the retrieval models: $78.7 \to 78.7$, $86.2 \to 86.3$ and $89.7 \to 89.7$ (\cref{app:squad}).

\subsection{Linear
patches do not provide any additional knowledge}

The linear patch is apparently needed and helps with the removal of the retrieval augmentation -- however, it is not acceptable to use a patch, which is doing more than `patching' and which adds some additional knowledge to the language model. This might even invalidate the positive results of retrieval-augmented pretraining on syntactic understanding. We will therefore focus on these tasks in this section. 

We can test if the patch provides additional knowledge by examining models that work well without it -- for them, patching should essentially be a no-operation that does not boost the performance. In our case, the models pretrained without any retrieval are the ones that do not need patching -- as they never use cross-attention. Looking at the \textsc{x-small}, \textsc{small} and \textsc{base} no-retrieval model, we can see that adding the linear patch does not lead to a better performance on linear probing: with the LAS scores $63.3\to63.4$, $71.2\to69.9$ and $78.1\to77.9$ (\cref{tab:linear-probing}). The same applies for the average BLiMP results: $73.4\to73.2$, $78.6\to78.6$ and $82.9\to82.8$ (\cref{tab:blimp}); as well as for the average MSGS results: $-0.55\to-0.57$, $-0.52\to-0.56$ and $-0.47\to-0.40$ (\cref{tab:MSGS}). The last result is the only exception, but we believe that it might be caused by the high variation of the MSGS results (as visible in \cref{fig:MSGS}). In addition, the trend applied for the world knowledge and language understanding tasks -- linear patching does not give a consistent advantage to the `no-retrieval' model. We therefore conclude that the separated language model do not gain an unfair advantage by using linear patching.


\section{Pretraining details}
\label{app:pretraining}

We pretrained a number of masked language models on a relatively small dataset of about 400 million words. That is why we follow the optimized LTG-BERT training recipe from \newcite{samuel-etal-2023-trained}, which showed to be effective for a low-resource setting.

We use WordPiece as the subword tokenizer \citep{https://doi.org/10.48550/arxiv.1609.08144} and set its vocabulary size to 16\,384, following LTG-BERT. We represent the text as a sequence of UTF-8 bytes instead of Unicode characters, as proposed by \newcite{radford2019language}.

The training time is sped up by parallelization over multiple GPUs. The computationally most expensive models are the \textsc{base}-sized retrieval-augmented models, these are pretrained on 128 AMD MI250X GPUs for 414 minutes. All the experiments were run on the LUMI supercomputer.\footnote{\url{https://www.lumi-supercomputer.eu/sustainable-future/}}.

\vspace{1em}

\begin{table*}[h!]
\centering
\small
\begin{tabular}{@{}lc@{}}
\toprule
\textbf{Hyperparameter} & \textsc{x-small} / \textsc{small} / \textsc{base} \\ \midrule
Number of layers        & 12 / 12 / 12          \\
Hidden size             & 192 / 384 / 768          \\
FF intermediate size    & 512 / 1\,024 / 2\,048    \\
Vocabulary size         & 16\,384         \\
Attention heads         & 3 / 6 / 12           \\
Dropout                 & 0.1           \\
Attention dropout       & 0.1           \\
Training steps          & 15\,625       \\
Batch size              & 32\,768       \\
Sequence length         & 128        \\
Warmup steps            & 250 (1.6\% steps)         \\
Initial learning rate   & 0.01          \\
Final learning rate     & 0.001          \\
Learning rate decay     & cosine        \\
Weight decay            & 0.1          \\
Layer norm $\epsilon$   & 1e-7          \\
Optimizer               & LAMB         \\
LAMB $\epsilon$         & 1e-6          \\
LAMB $\beta_1$          & 0.9           \\
LAMB $\beta_2$          & 0.98          \\
Gradient clipping       & 2.0           \\ \bottomrule
\end{tabular} %
\caption{Pre-training hyperparameters for all three model sizes. The retrieval and no-retrieval models use the same hyperparameters.}
\label{tab:hyperparams}
\end{table*}


\section{Evaluation details}
\label{app:evaluation}

\subsection{LAMA probing}
\label{app:lama}

We calculate rank-based metrics for all subsets: mean precision at $k$ (P@$k$) and mean reciprocal rank (MRR). For a given statement, we count a fact as correctly predicted if the object is ranked among the top k results, and wrong otherwise. As the models are trained on a relatively small corpus in a narrow domain, the vocabulary is smaller than a typical language model. To account for this during evaluation, we remove all statements where the correct token is not in the models' vocabularies.

Both baselines and models trained with retrieval have the same vocabulary, so we do not need to account for differences between the two with respect to OOV words. However, as our models are trained only on a subset of Wikipedia, the proportion of OOV words with respect to the gold tokens in the LAMA probe is significant. We account for this by removing all statements where the correct token is not in the models' vocabularies. \cref{tab:appendix_lama_stats} shows the number of original statements and how many were included in the evaluations.


\begin{table}[h]
\centering
\small
    \begin{tabular}{@{}l@{\hspace{2em}}rr@{}}
    \toprule
    \textbf{Dataset} & \textbf{\# Facts} & \textbf{\# Facts evaluated on} \\
    \midrule
    SQuAD & $305$ & $221$ \\
    ConceptNet & $29\,774$ & $16\,997$ \\
    TREx & $34\,039$& $22\,550$ \\
    \bottomrule
    \end{tabular}
    \caption{Statistics about the number of facts in the different subsets of LAMA \citep{petroni-etal-2019-language}}
    \label{tab:appendix_lama_stats}
\end{table}

\begin{table*}[!h]
\resizebox{\textwidth}{!}{%
\begin{NiceTabular}{@{}lrrrr@{\hspace{3em}}rrrr@{\hspace{3em}}rrrr@{}}
\toprule
                    \Block{2-1}{\textbf{Model}} & \Block{1-4}{\textbf{ConceptNet}} & & & & \Block{1-4}{\textbf{SQuAD}} & & & &\Block{1-4}{\textbf{TREx}}\\
                    & P@1 & P@10 &P@100 & MRR & P@1 & P@10 & P@100 & MRR & P@1 & P@10 & P@100 & MRR \\ 
\midrule

    \textsc{reference model}& & & \\
 \hspace{1em}\textit{bert-base-cased}& \textit{17.20} & \textit{44.31} & \textit{70.59} & \textit{26.00} & \textit{21.71} & \textit{65.15} & \textit{79.63} & \textit{34.00} & \textit{52.55} & \textit{80.08} & \textit{92.27} & \textit{62.00} \\[0.5em]
 
 \textsc{base}& & & \\
 \hspace{1em}$\bm{-}$ retrieval pretraining (patch) & \underline{12.97} & \textbf{37.46} & \textbf{60.15} & \textbf{20.48} & \textbf{21.71} & \textbf{65.15} & \textbf{72.39} & \textbf{31.98} & \textbf{43.31} & \underline{75.11} & \textbf{88.72} & \textbf{53.84} \\
 \hspace{1em}$\bm{-}$ retrieval pretraining (no patch) & \textbf{13.03} & \underline{36.62} & \underline{60.06} & \underline{20.34} & \underline{21.17} & \textbf{65.15} & \textbf{72.39} & \underline{32.09} & \underline{42.82} & \textbf{75.11} & \underline{88.67} & \underline{53.62} \\
 \hspace{1em}$\bm{+}$ retrieval pretraining (50\% noise, patch) & 10.80 & 33.51 & 56.63 & 17.74 & 14.47 & \underline{43.43} & \underline{65.15} & 23.15 & 37.38 & 72.92 & 87.91 & 49.09 \\
 \hspace{1em}$\bm{+}$ retrieval pretraining (25\% noise, patch) & 11.16 & 31.72 & 56.78 & 18.08 & 14.47 & 36.19 & \textbf{72.39} & 23.44 & 36.26 & 72.76 & 87.15 & 48.29 \\
 \hspace{1em}$\bm{+}$ retrieval pretraining (0\% noise, patch) & 9.30 & 27.81 & 54.71 & 14.93 & 7.23 & \underline{43.43} & \textbf{72.39} & 15.75 & 29.62 & 66.08 & 85.77 & 41.51 \\
 \hspace{1em}$\bm{+}$ retrieval pretraining (0\% noise, no patch) & 5.54 & 19.35 & 40.99 & 9.78 & 7.23 & 14.47 & 50.67 & 10.50 & 20.41 & 55.09 & 79.13 & 31.42 \\[0.5em]

 \textsc{small}& & &\\
\hspace{1em}$\bm{-}$ retrieval pretraining (patch) & \underline{10.24} & \underline{29.89} & \underline{54.04} & \underline{16.64} & \underline{14.47} & \textbf{57.91} & \underline{72.39} & \underline{25.59} & \textbf{37.13} & \underline{68.86} & \textbf{86.36} & \textbf{47.62} \\
\hspace{1em}$\bm{-}$ retrieval pretraining (no patch) & \textbf{10.90} & \textbf{30.42} & \textbf{54.89} & \textbf{17.25} & \textbf{21.71} & \underline{50.67} & \textbf{79.63} & \textbf{28.29} & \underline{36.77} & \textbf{69.19} & \underline{85.97} & \underline{47.44} \\
\hspace{1em}$\bm{+}$ retrieval pretraining (0\% noise, patch) & 6.57 & 22.77 & 48.51 & 11.77 & 7.23 & 28.95 & 65.15 & 15.38 & 25.71 & 58.71 & 81.47 & 36.31 \\
\hspace{1em}$\bm{+}$ retrieval pretraining (0\% noise, no patch) & 1.21 & 5.83 & 18.52 & 2.72 & 0.0 & 7.23 & 21.17 & 3.92 & 5.58 & 15.44 & 34.48 & 8.88 \\[0.5em]

 \textsc{x-small}& & &\\
\hspace{1em}$\bm{-}$ retrieval pretraining (patch) & \textbf{5.82} & \textbf{21.52} & \underline{45.03} & \textbf{10.67} & \textbf{7.23} & \underline{36.19} & \underline{65.15} & \underline{14.57} & \underline{27.44} & \underline{61.10} & \underline{83.13} & \underline{38.48} \\
\hspace{1em}$\bm{-}$ retrieval pretraining (no patch) & \underline{5.26} & \underline{21.33} & \textbf{45.60} & \underline{9.91} & \textbf{7.23} & \textbf{43.43} & \textbf{72.39} & \textbf{14.74} & \textbf{27.92} & \textbf{61.11} & \textbf{83.45} & \textbf{39.17} \\
\hspace{1em}$\bm{+}$ retrieval pretraining (0\% noise, patch) & 4.3 & 14.80 & 37.45 & 7.47 & \textbf{7.23} & 14.47 & 57.91 & 10.64 & 14.03 & 45.12 & 73.80 & 23.42 \\ 
\hspace{1em}$\bm{+}$ retrieval pretraining (0\% noise, no patch) & 0.0 & 0.0 & 1.95 & 0.0 & \underline{0.0} & 0.0 & 0.0 & 0.0 & 0.0 & 0.0 & 0.0 & 0.0 \\

\bottomrule
\end{NiceTabular}%
}
\caption{Results on zero-shot evaluation on different subsets of the LAMA probe. MRR is calculated at $k=100$. The \textbf{bold} numbers represent the best model at each size, while the \underline{underline} is the second best.} 
\label{tab:lama}
\end{table*}


\subsection{Linear probing}
\label{app:lp}

With linear probing, we are measuring how much information about a downstream task can be extracted from the hidden representations with a simple linear transformation. The reasoning is that a model with a better syntactic understanding should encode more of the syntactic information in the latent vectors. However, note that the results also depend on the accessibility of the syntactic information, because we do not use any nonlinear transformations. The reason for avoiding non-linearities is that we want to measure the amount of knowledge already stored in the language model, not the knowledge learned by the complex nonlinear transformation.

In order to parse an input, we first extract subword representations $s_{i,k}$ from a frozen language model, for all positions $i$ and layers $k$. To get a vector representation $h_t$ for the $t^\textrm{th}$ word-span, we apply two pooling operations on the subword-token representations $s_{t,k}$: first, we pool the vectors at all layers by taking a learned convex combination:
\[\hat{s}_t = \sum_{k=1}^{L}{\gamma_k s_{t,k}},\]
where $\gamma_k \in \mathbb{R}$ (based on the observation that the syntactic information is present more strongly in some layers \citep{kondratyuk-straka-2019-75, rogers-etal-2020-primer}, we allow the model to select the most useful combination of layers). Next, since one word-span can be split into multiple subwords, we average the respective subword representation and get the final contextualized representation $h_t$. 

Finally, to predict the dependency tree, we take a similar approach to \newcite{dozat2017deep} and employ a \textit{shallow} bilinear attention mechanism -- without any nonlinear activations. The logit of an arc between words at positions $i$ and $j$ is defined as:
\[\text{arc}_{i\to j} = h_i U h_j + h_i u_{\text{head}} + h_j u_{\text{dep}} + b,\]
where $U, u_{\text{head}}, u_{\text{dep}}$ and $b$ are learnable parameters; the original parameters of the language model remain frozen. Then we apply the Chu-Liu-Edmonds maximum spanning tree algorithm on the directed graph of arc probabilities \citep{1570854175817997952}. The edge-label prediction also follows \newcite{dozat2017deep} in a similar manner.

We use the gold standard dependency corpus for English \citep{silveira-etal-2014-gold}, specifically its conversion to Universal Dependencies 2.12 \citep{nivre-etal-2017-universal}.\footnote{Available online: \url{https://github.com/UniversalDependencies/UD_English-EWT}.}

\paragraph{Significance test} In order to test the statistical significance of the improvement by retrieval pretraining, we use the Almost Stochastic Order test \citep[ASO;][]{del2018optimal, dror2019deep} implemented by \citet{ulmer2022deep}. We compare the \textit{`$\bm{-}$ retrieval pretraining (no patch)'} results with \textit{`$\bm{+}$ retrieval pretraining (patch)'} results (\cref{tab:linear-probing}). All models were finetuned on five random seeds and we use ASO with a confidence level of $\alpha = 0.05$. The almost stochastic dominance ($\epsilon_\text{min} < \tau$ with $\tau = 0.2$) on the primary LAS metric is achieved by all three sizes of models,\footnote{This is clearly achieved with $\epsilon_\text{min}$ of $0.0$, $0.0$ and $0.0016$ for the \textsc{base}, \textsc{small} and \textsc{x-small} sizes, respectively.} which shows that \textbf{the improvement is statistically significant}.

\begin{table}[!ht]
\small
  \centering
    \begin{tabular}{@{}lccc@{}}
    \toprule
    \textbf{Model} & \textbf{UAS} & \textbf{LAS} & \textbf{CLAS} \\
    \midrule
    \textsc{reference model}& & & \\
 \hspace{1em}\textit{bert-base-cased}& \textit{85.01}$^{\pm0.08}$ &	\textit{81.96}$^{\pm0.11}$ &	\textit{77.98}$^{\pm0.16}$ \\[0.5em]
 \textsc{base}& & & \\
  \hspace{1em}$\bm{-}$ retrieval pretraining (patch) &81.19$^{\pm0.09}$ &	77.90$^{\pm0.07}$ &	73.93$^{\pm0.11}$ \\
  \hspace{1em}$\bm{-}$ retrieval pretraining (no patch) & 81.42$^{\pm0.08}$ & 78.06$^{\pm0.09}$ &	74.14$^{\pm0.11}$ \\
 \hspace{1em}$\bm{+}$ retrieval pretraining (50\% noise, patch) & 82.95$^{\pm0.12}$ & 79.82$^{\pm0.10}$ & 76.18$^{\pm0.09}$ \\
 \hspace{1em}$\bm{+}$ retrieval pretraining (25\% noise, patch) & \underline{83.06}$^{\pm0.08}$ &	\underline{79.94}$^{\pm0.12}$ &	\underline{76.46}$^{\pm0.15}$ \\
\hspace{1em}$\bm{+}$ retrieval pretraining (0\% noise, patch) & \textbf{83.41}$^{\pm0.09}$ & \textbf{80.25}$^{\pm0.11}$	& \textbf{76.72}$^{\pm0.17}$ \\
 \hspace{1em}$\bm{+}$ retrieval pretraining (0\% noise, no patch) & 81.28$^{\pm0.08}$ &	78.07$^{\pm0.07}$ &	74.17$^{\pm0.14}$ \\[0.5em]
 \textsc{small}& & &\\
\hspace{1em}$\bm{-}$ retrieval pretraining (patch) & 73.15$^{\pm0.02}$ &	69.93$^{\pm0.01}$ &	64.63$^{\pm0.05}$  \\
\hspace{1em}$\bm{-}$ retrieval pretraining (no patch) & \underline{74.34}$^{\pm0.09}$ & \underline{71.17}$^{\pm0.11}$ &	\underline{66.03}$^{\pm0.19}$ \\
\hspace{1em}$\bm{+}$ retrieval pretraining (patch) & \textbf{74.91}$^{\pm0.07}$ &	\textbf{71.72}$^{\pm0.12}$ &	\textbf{66.40}$^{\pm0.17}$ \\
\hspace{1em}$\bm{+}$ retrieval pretraining (no patch) & 67.86$^{\pm0.07}$ &	64.57$^{\pm0.09}$ &	58.25$^{\pm0.11}$ \\[0.5em]
 \textsc{x-small}& & &\\
\hspace{1em}$\bm{-}$ retrieval pretraining (patch) & \underline{67.24}$^{\pm0.03}$ &	\underline{63.41}$^{\pm0.05}$ &	\textbf{57.01}$^{\pm0.11}$ \\
\hspace{1em}$\bm{-}$ retrieval pretraining (no patch) & 67.13$^{\pm0.07}$ &	63.31$^{\pm0.07}$ &	56.86$^{\pm0.13}$ \\
\hspace{1em}$\bm{+}$ retrieval pretraining (patch) & \textbf{67.46}$^{\pm0.18}$ &	\textbf{63.61}$^{\pm0.13}$ &	\underline{56.96}$^{\pm0.15}$ \\
\hspace{1em}$\bm{+}$ retrieval pretraining (no patch) &  50.26$^{\pm0.08}$ &	46.23$^{\pm0.08}$ &	40.51$^{\pm0.18}$ \\
   \bottomrule 
    \end{tabular}
    \caption{The results of linear probing for dependency parsing. We evaluate the predictions with three standard metric: the unlabeled attachment score (UAS), the labeled attachment score (LAS) and the content-word labeled attachment score \citep[CLAS;][]{nivre-fang-2017-universal}.The \textbf{bold} numbers represent the best model at each size, while the \underline{underline} is the second best.}
    \label{tab:linear-probing}
\end{table}


\begin{figure*}[t!]
    \centering
    \begin{subfigure}[t]{0.49\textwidth}
        \centering
        \includegraphics[width=\textwidth]{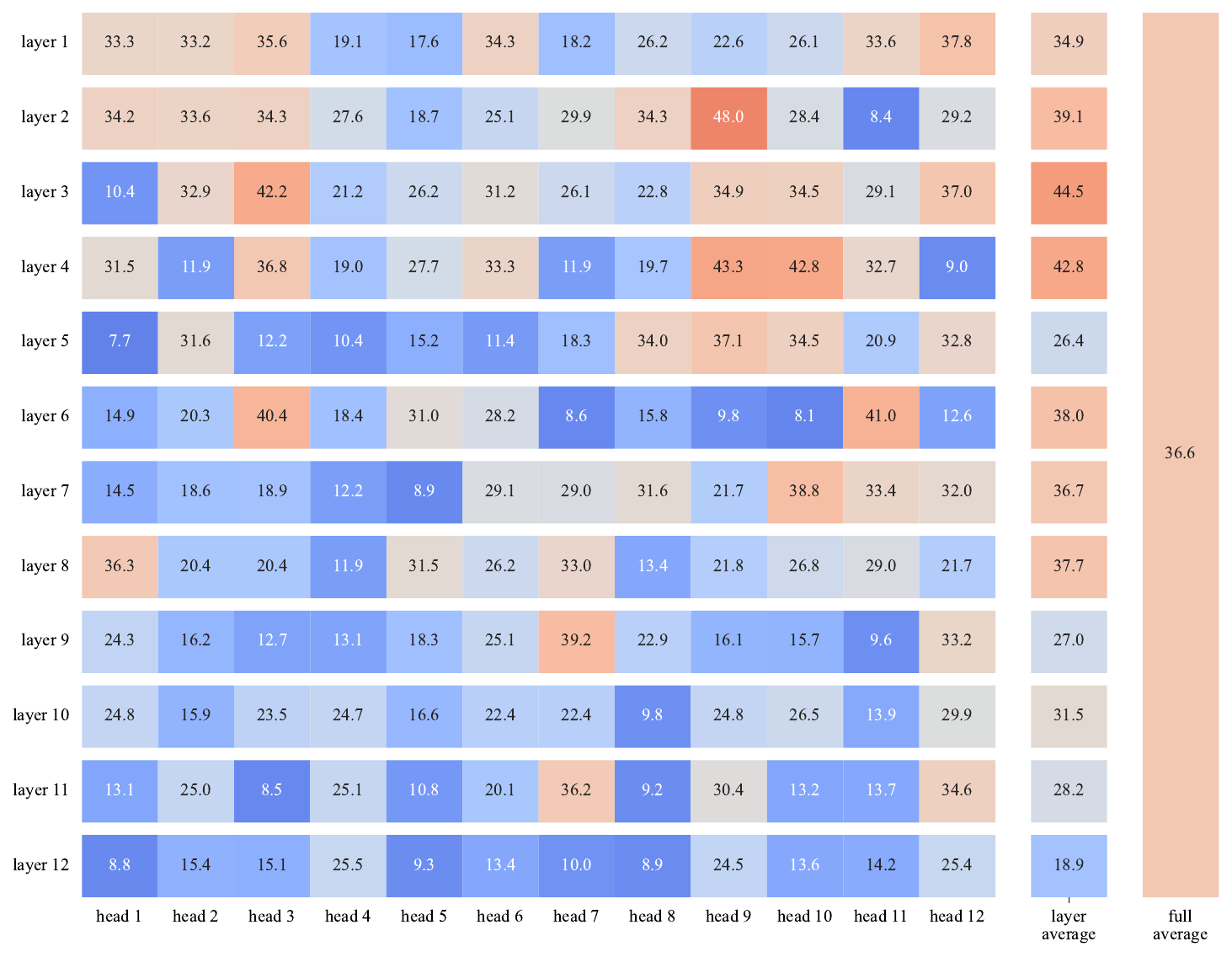}
        \caption{\textsc{base}: no retrieval pretraining.}
        \label{fig:uuas-1}
    \end{subfigure}%
    ~~
    \begin{subfigure}[t]{0.49\textwidth}
        \centering
        \includegraphics[width=\textwidth]{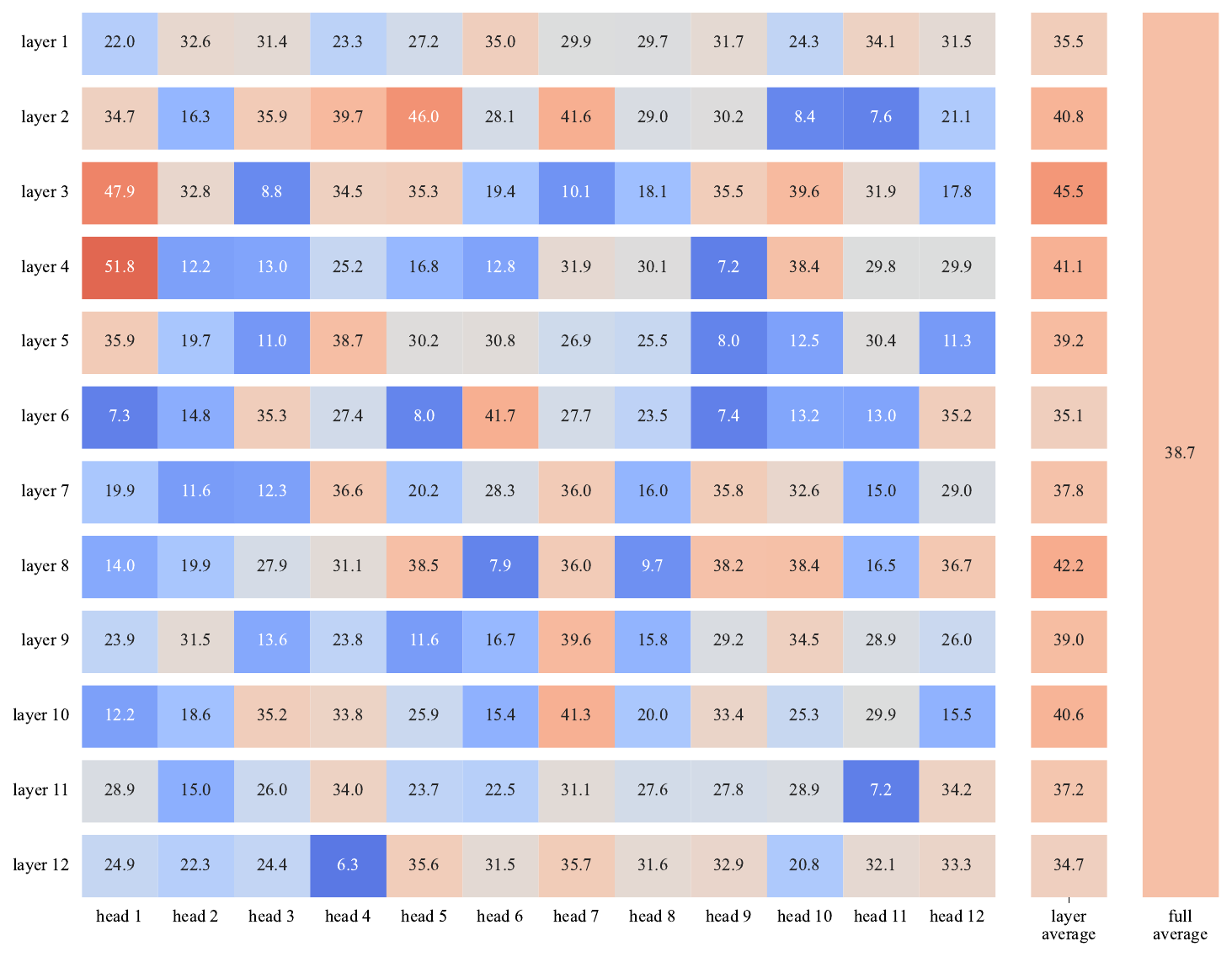}
        \caption{\textsc{base}: retrieval-augmented pretraining.}
    \end{subfigure}

    \vspace{2em}

    \begin{subfigure}[t]{0.49\textwidth}
        \centering
        \includegraphics[width=\textwidth]{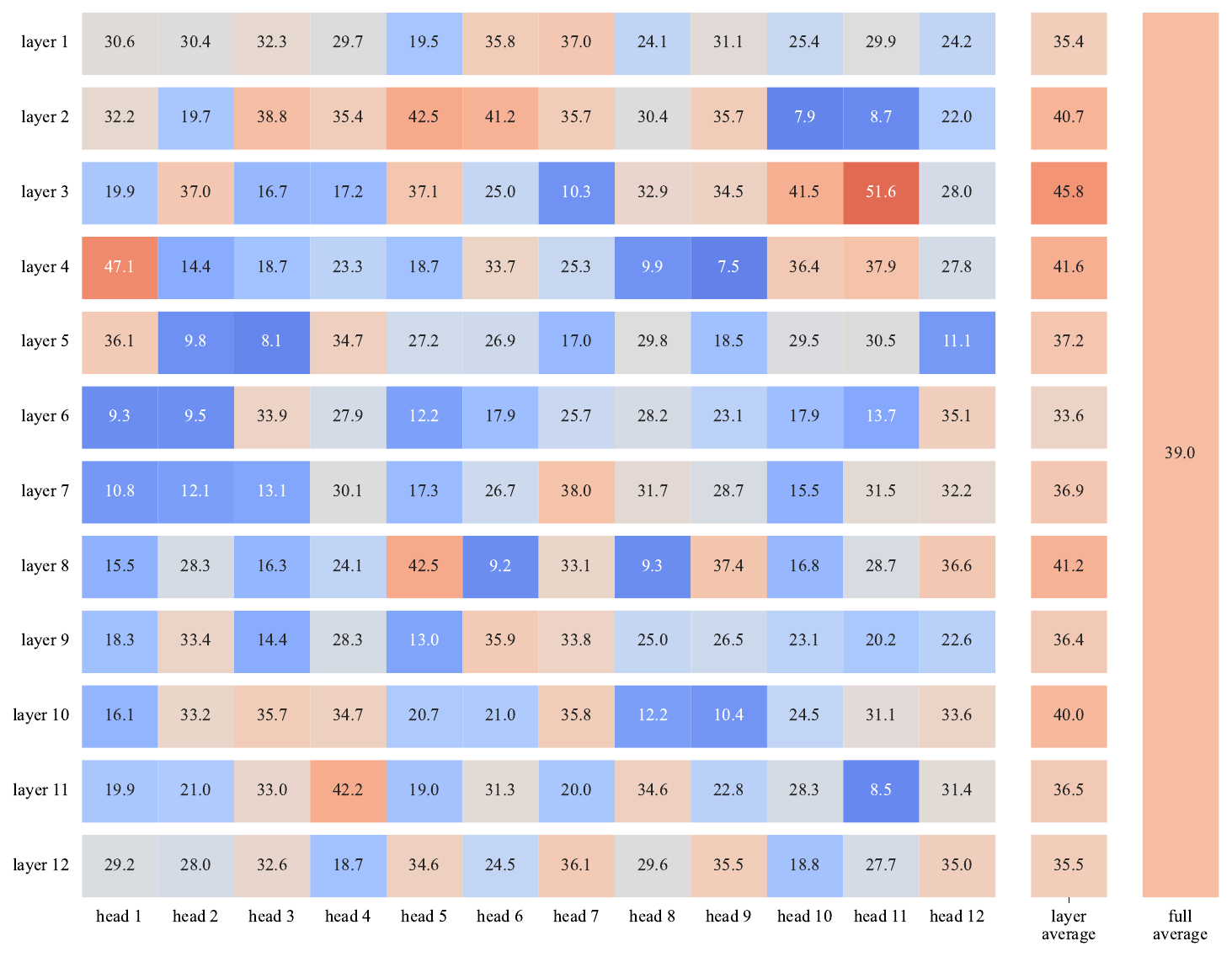}
        \caption{\textsc{base}: retrieval-augmented pretraining with 25\% noise.}
    \end{subfigure}%
    ~~
    \begin{subfigure}[t]{0.49\textwidth}
        \centering
        \includegraphics[width=\textwidth]{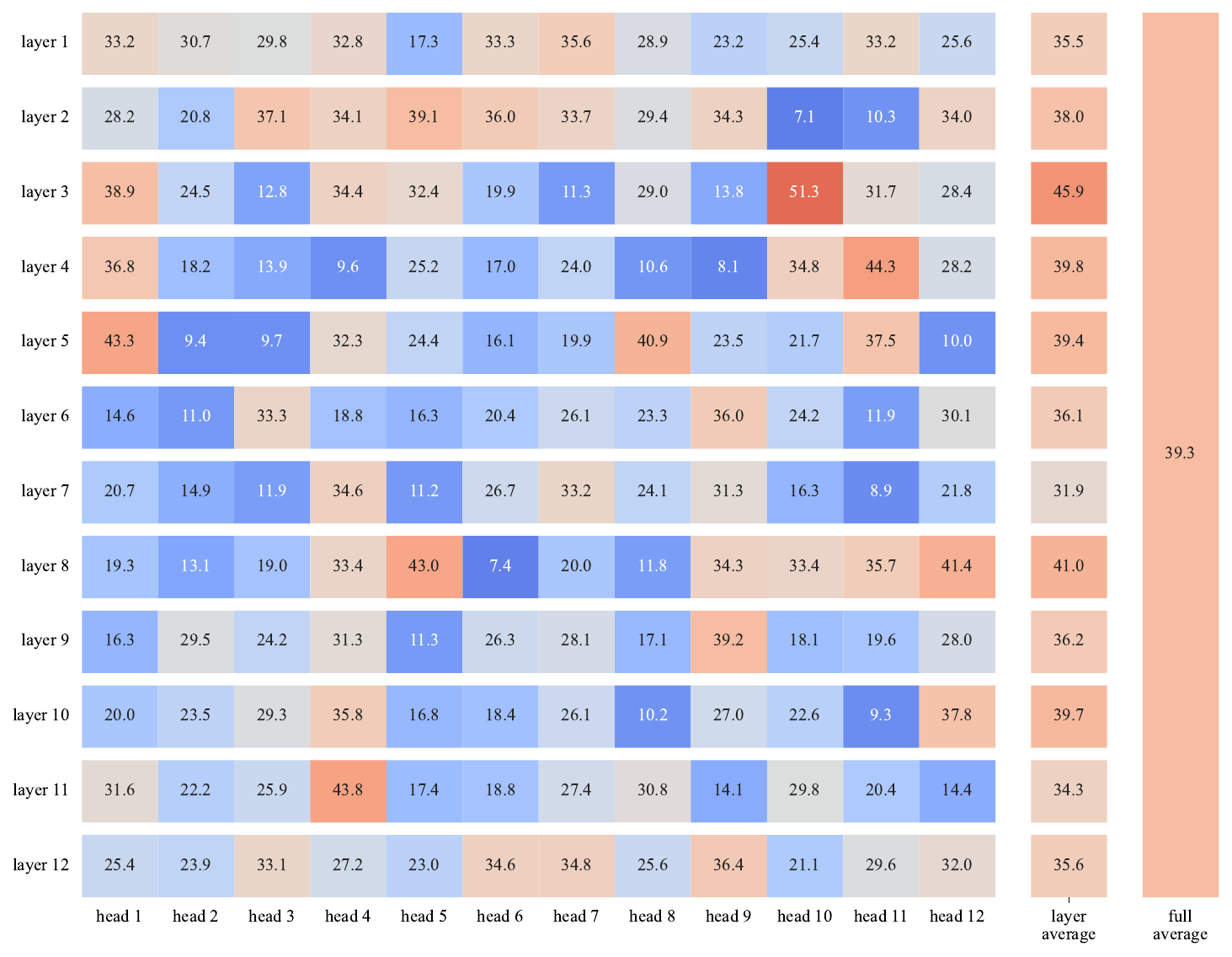}
        \caption{\textsc{base}: retrieval-augmented pretraining with 50\% noise.}
    \end{subfigure}

    \vspace{2em}

    \begin{subfigure}[t]{0.272\textwidth}
        \centering
        \includegraphics[width=\textwidth]{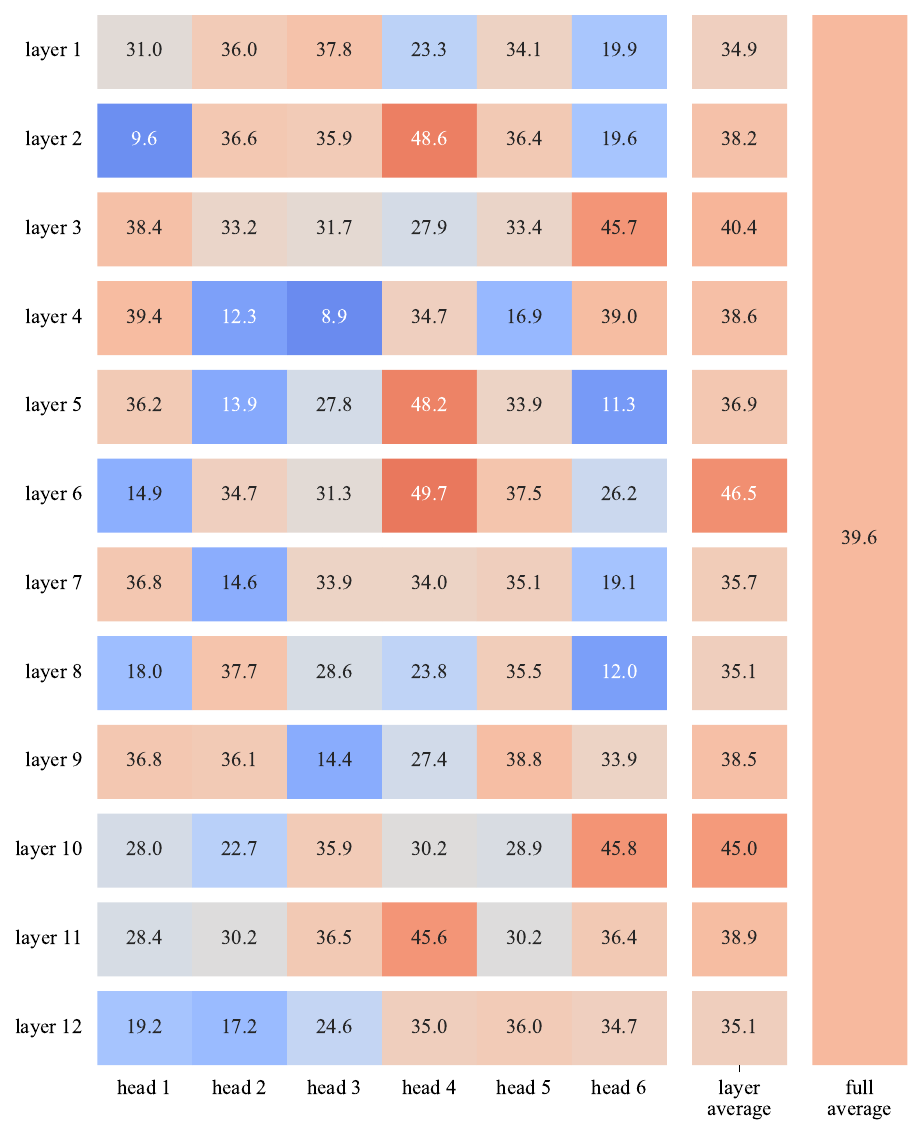}
        \caption{\textsc{small}: no retrieval pretraining.}
    \end{subfigure}%
    ~~ 
    \begin{subfigure}[t]{0.272\textwidth}
        \centering
        \includegraphics[width=\textwidth]{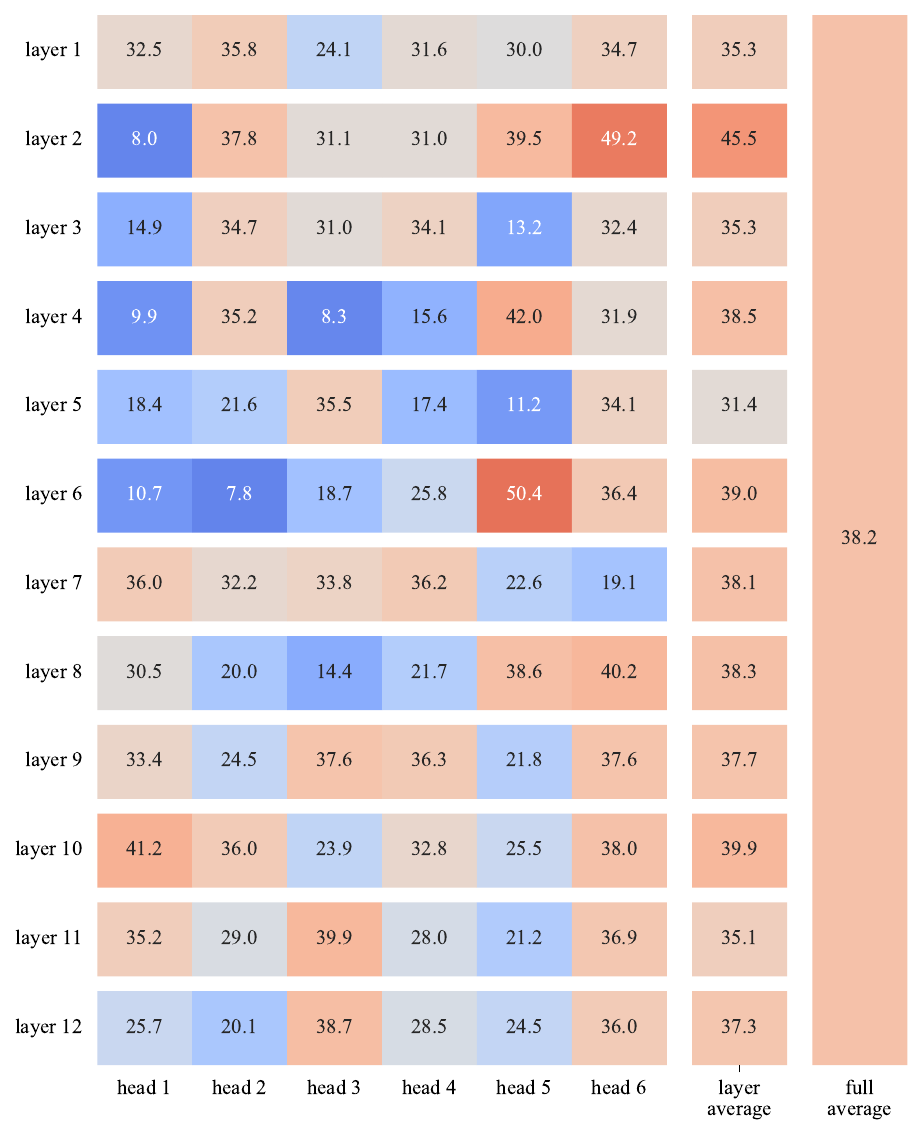}
        \caption{\textsc{small}: retrieval-augmented pretraining.}
    \end{subfigure}
    ~~
    \begin{subfigure}[t]{0.193\textwidth}
        \centering
        \includegraphics[width=\textwidth]{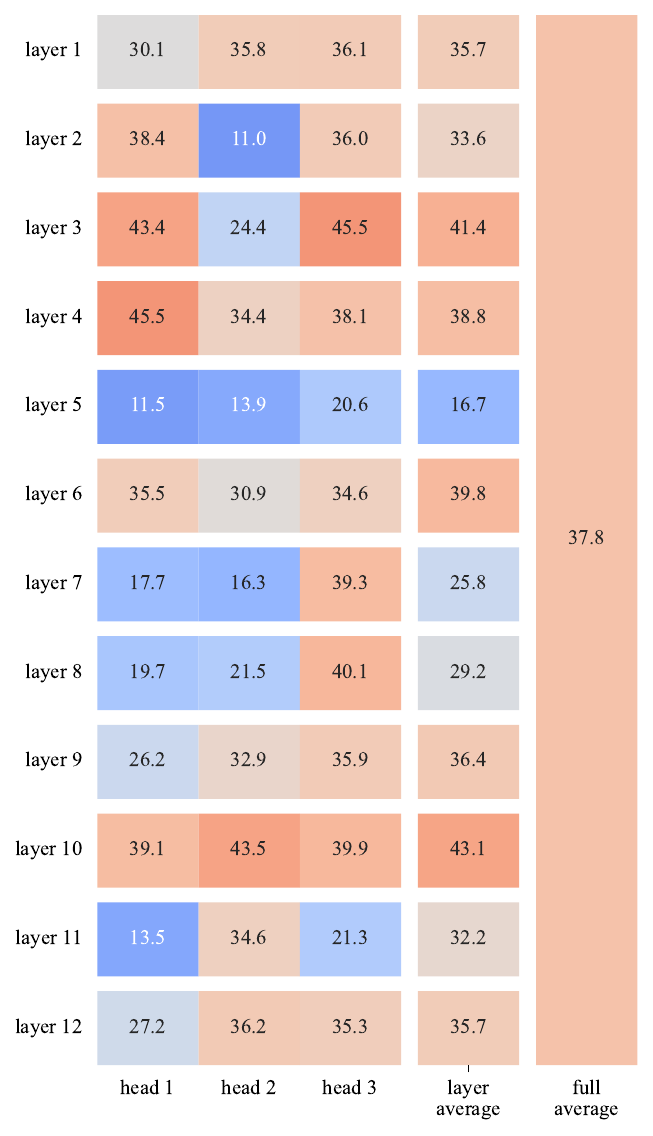}
        \caption{\textsc{x-small}: no retrieval pretraining.}
    \end{subfigure}
    ~~
    \begin{subfigure}[t]{0.193\textwidth}
        \centering
        \includegraphics[width=\textwidth]{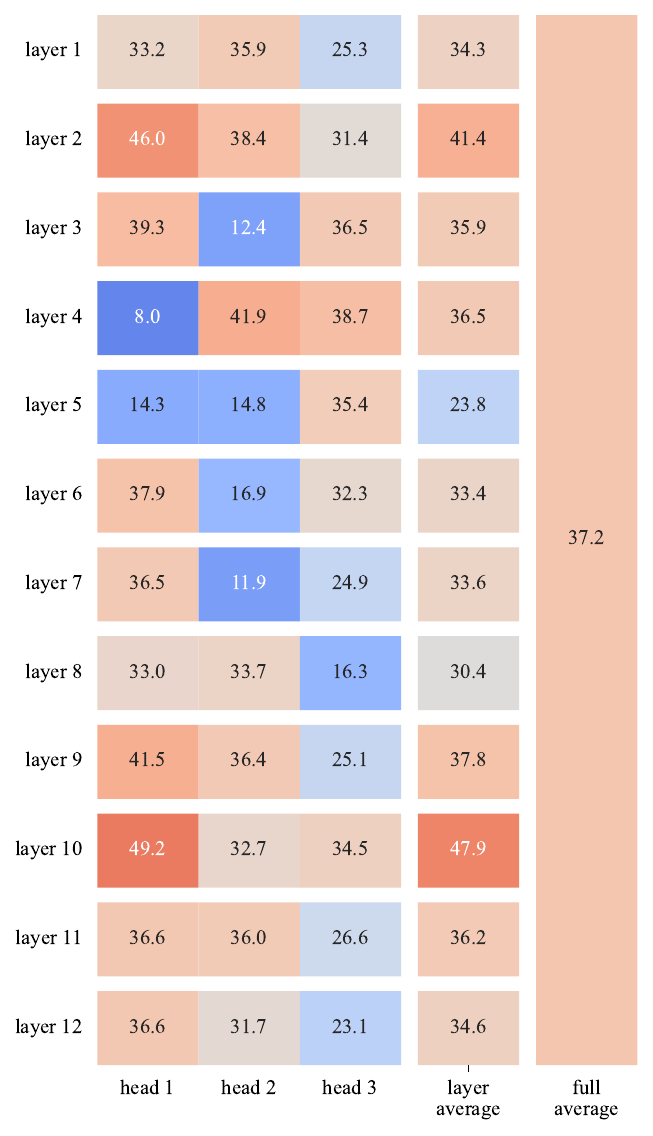}
        \caption{\textsc{x-small}: retrieval-augmented pretraining.}
    \end{subfigure}
    \caption{The undirected unlabeled attachment scores (UUAS) of attention probing with every head and layer combination. The plot also shows the UUAS scores of attention matrices averaged across each layer and across the whole language model.}
    \label{fig:attention-probe}
\end{figure*}


\subsection{Attention probing}
\label{app:ap}

We mostly follow \newcite{raganato-tiedemann-2018-analysis}, and \newcite{ravishankar-etal-2021-attention} in their evaluation setup of attention probing. Our goal is to decode dependency trees directly from the attention weights -- with the idea that a language model with better syntactic understanding should better utilize the hierarchical syntactic structure in its attention mechanism.

First, given a sentence of length $T$, we pass it through the language model and separately save the attention probabilities $A_{\ell,h} \in \mathbb{R}^{T\times T}$ for every layer $\ell$ and attention head $h$. To get elements that correspond to the surface words (not the tokenized subwords), we remove the rows and columns that correspond to the special \texttt{[CLS]} and \texttt{[SEP]} tokens, and we take the sum of the columns and the mean of the rows that correspond to one word split into multiple subwords. Then we make the attention matrix symmetric by multiplying it element-wise with its transpose: $A_{\ell,h} \leftarrow A_{\ell,h} \cdot A_{\ell,h}^\intercal$. Finally, we interpret $A_{\ell,h}$ as the weighted adjacency matrix of a fully-connected undirected graph and extract the dependency tree by finding the maximum spanning tree of that graph \citep{zbMATH02560699}. The fitness the decoded graph is then measured via the undirected unlabeled attachment score \citep[UUAS;][]{osti_10233691}.

As per \newcite{ravishankar-etal-2021-attention}, we report the best head score as the primary metric in \cref{tab:results}. However, fine-grained results for all heads are given in \cref{fig:attention-probe}.



\subsection{BLiMP}
\label{app:blimp}

The Benchmark of Linguistic Minimal Pairs for English \citep{warstadt-etal-2020-blimp-benchmark} attempts to measure the linguistic knowledge of a language model in a zero-shot manner -- without any additional training. It consists of 67 tasks, each focuses on a specific linguistic feature, which is tested with 1\,000 automatically generated sentence pairs. Each pair of sentences differs minimally on the surface level, but only one of the sentences is grammatically valid. The tasks are clustered into the following subgroups, with descriptions taken from \citet{warstadt-etal-2020-blimp-benchmark}:
\begin{itemize}
    \item \textsc{Anaphor Agreement} (AA): the requirement that reflexive pronouns like \textit{herself} (also known as anaphora) agree with their antecedents in person, number, gender, and animacy.
    \item \textsc{Argument structure} (AS): the ability of different verbs to appear with different types of arguments. For instance, different verbs can appear with a direct object, participate in the causative alternation, or take an inanimate argument.
    \item \textsc{Binding} (B): the structural relationship between a pronoun and its antecedent.
    \item \textsc{Control/raising} (CR): syntactic and semantic differences between various types of predicates that embed an infinitival VP. This includes control, raising, and \textit{tough}-movement predicates.
    \item \textsc{Determiner-noun agreement} (DNA): number agreement between demonstrative determiners (e.g., \textit{this/these}) and the associated noun.
    \item \textsc{Ellipsis} (E): the possibility of omitting expressions from a sentence. Because this is difficult to illustrate with sentences of equal length, our paradigms cover only special cases of noun phrase ellipsis that meet this constraint.
    \item \textsc{Filler-gap} (FG): dependencies arising from phrasal movement in, for example, \textit{wh}-questions.
    \item \textsc{Irregular forms} (IF): irregular morphology on English past participles (e.g., \textit{awoken}).
    \item \textsc{Island effects} (IE): restrictions on syntactic environments where the gap in a filler-gap dependency may occur.
    \item \textsc{NPI licensing} (NL): restrictions on the distribution of \textit{negative polarity items} like \textit{any} and \textit{ever} limited to, for example, the scope of negation and \textit{only}.
    \item \textsc{Quantifiers} (Q): restrictions on the distribution of quantifiers.  Two such restrictions are covered: superlative quantifiers (e.g., \textit{at least}) cannot be embedded under negation, and definite quantifiers and determiners cannot be subjects in existential-\textit{there} constructions.
    \item \textsc{Subject-verb agreement} (SVA): subjects and present tense verbs must agree in number.
\end{itemize}

We use the intrinsic ability of language models to estimate the probability of any text segment, and measure how often the evaluated language model assigns a higher probability to the grammatically correct sentence. Specifically we employ the \textit{pseudo-log-likelihood score} by \newcite{wang-cho-2019-bert} and \newcite{salazar-etal-2020-masked} to rank the sentences with a masked language model. We also follow the observation by \newcite[][Appendix A]{samuel-2023-mean} that the results on BLiMP greatly depend on temperature scaling -- to do a fair comparison between different types of language models, they proposed to search for the optimal temperature value for each evaluated model.

\cref{tab:blimp} shows the detailed results of each model for each subgroup mentioned above. At all sizes, we observe that retrieval pre-trained models perform better with quantifiers and binding.

\begin{table*}[!h]
\resizebox{\textwidth}{!}{%
\begin{tabular}{@{}lrrrrrrrrrrrrr@{}}
\toprule
                   \textbf{Model} & \textbf{AS} & \textbf{Q} & \textbf{IF} & \textbf{FGD} & \textbf{IE} & \textbf{AA} & \textbf{NL} & \textbf{SVA} & \textbf{E} & \textbf{B} & \textbf{CR} & \textbf{DNA} & \textbf{Average} \\ 
\midrule
    \textsc{reference model}& & & \\
 \hspace{1em}\textit{bert-base-cased}& \textit{86.22} & \textit{60.80} & \textit{97.95} & \textit{87.49} & \textit{71.79} & \textit{97.45} & \textit{86.50} & \textit{94.53} & \textit{89.80} & \textit{82.20} & \textit{85.58} & \textit{97.56} & \textit{85.56} \\[0.5em]
 
 \textsc{base}& & & \\
 \hspace{1em}$\bm{-}$ retrieval pretraining (patch) & 81.97 & 65.85 & 95.35 & 86.50 & 65.86 & 97.90 & \underline{84.77} & \underline{94.57} & \textbf{91.75} & 72.77 & 79.52 & 96.76  & 82.77 \\
\hspace{1em}$\bm{-}$ retrieval pretraining (no patch) & 82.14 & 65.90 & \underline{95.50} & 86.59 & 66.39 & 97.85 & \textbf{84.89} & 94.17 & \underline{91.65} & \underline{73.10} & 79.26 & 96.85 & \underline{82.87} \\
\hspace{1em}$\bm{+}$ retrieval pretraining (50\% noise, patch) & 81.26 & 62.25 & 94.40 & 85.84 & 63.76 & \textbf{98.40} & 80.49 & 93.57 & 89.40 & 70.40 & 79.80 & 96.94 & 81.31 \\
\hspace{1em}$\bm{+}$ retrieval pretraining (25\% noise, patch) & \underline{82.67} & 65.33 & 94.30 & \underline{87.33} & \textbf{68.73} & \underline{98.10} & 82.97 & 93.38 & 89.20 & 69.63 & \textbf{81.72} & \underline{97.09} & 82.74 \\
\hspace{1em}$\bm{+}$ retrieval pretraining (0\% noise, patch) & \textbf{82.99} & \textbf{68.70} & \textbf{95.65} & \textbf{87.81} & \underline{67.70} & 96.50 & 83.11 & \textbf{95.35} & 90.45 & 69.33 & \underline{81.68} & \textbf{97.55} & \textbf{83.15} \\
  \hspace{1em}$\bm{+}$ retrieval pretraining (0\% noise, no patch) & 79.28 & \underline{68.45} & 90.25 & 86.89 & 66.03 & 92.30 & 74.10 & 89.22 & 88.70 & \textbf{74.20} & 79.88 & 95.78  & 80.67  \\ [0.5em]

 \textsc{small}& & &\\
\hspace{1em}$\bm{-}$ retrieval pretraining (patch) & \underline{78.99} & \underline{64.08} & \textbf{94.50} & 80.71 & \textbf{57.91} & \underline{96.75} & 74.87 & \textbf{91.78} & \underline{89.35} & 68.03 & \underline{77.86} & \textbf{95.95} & \underline{78.58} \\
\hspace{1em}$\bm{-}$ retrieval pretraining (no patch) & \textbf{79.50} & 62.50 & 92.70 & \textbf{82.41} & \underline{57.73} & \textbf{97.35} & 75.60 & 90.80 & 88.05 & 67.84 & 77.62 & \underline{95.94} & \underline{78.58} \\
\hspace{1em}$\bm{+}$ retrieval pretraining (0\% noise, patch) & 76.71 & 62.88 & \underline{93.45} & \underline{80.99} & 56.00 & 92.75 & \textbf{80.04} & \underline{91.07} & \textbf{90.90} & \underline{71.41} & \textbf{78.94} & 95.75 & \textbf{78.78} \\
\hspace{1em}$\bm{+}$ retrieval pretraining (0\% noise, no patch) & 69.87 & \textbf{68.70} & 89.50 & 74.66 & 49.51 & 89.75 & \underline{75.77} & 83.28 & 85.00 & \textbf{75.27} & 72.08 & 92.70 & 74.77 \\ [0.5em]

 \textsc{x-small}& & &\\
\hspace{1em}$\bm{-}$ retrieval pretraining (patch) &  71.22 & \underline{65.58} & \underline{93.25} & \underline{71.36} & 46.58 & \underline{93.70} & \underline{70.00} & \underline{87.75} & \textbf{86.75} & 68.03 & \underline{69.48} & \underline{92.54} & 73.18 \\
\hspace{1em}$\bm{-}$ retrieval pretraining (no patch) & \underline{72.17} & 64.60 & \textbf{94.30} & 70.96 & 44.95 & \textbf{93.75} & \textbf{70.19} & \textbf{88.45} & \underline{85.80} & \underline{69.04} & \textbf{70.26} & \textbf{93.34} & \textbf{73.36} \\
\hspace{1em}$\bm{+}$ retrieval pretraining (0\% noise, patch) & \textbf{72.22} & 64.08 & 90.10 & \textbf{74.30} & \underline{51.15} & 87.20 & 68.96 & 84.15 & 85.45 & \textbf{69.43} & 68.66 & 91.74 & \underline{73.31}\\
\hspace{1em}$\bm{+}$ retrieval pretraining (0\% noise, no patch) & 58.82 & \textbf{68.85} & 52.90 & 56.86 & \textbf{51.41} & 75.00 & 50.50 & 63.30 & 36.95 & 66.00 & 61.38 & 61.75 & 58.81 \\

\bottomrule
\end{tabular}%
}
\caption{Fine-grained BLiMP results. AS = argument structure, Q = quantifiers, IF = irregular forms, FGD = filler gap dependency, IE = island effects, AA = anaphor agreement, NL = NPI licensing, SVA = subject-verb agreement, E = ellipsis, B = binding, CR = control raising and DNA = determiner-noun agreement. The \textbf{bold} numbers represent the best model at each size, while the \underline{underline} is the second best.}
\label{tab:blimp}
\end{table*}

\subsection{MSGS}
\label{app:msgs}

The MSGS benchmark \citep{warstadt-etal-2020-learning} evaluates whether the model biases linguistic features or surface features. A score of 1 means only using the linguistic features, while a score of -1 is surface features only. To evaluate the performance we use the Mathews Correlation Coefficient (MCC), also called Linguistic Bias Score (LBS). The surface features in this dataset are (definitions taken from \citet{warstadt-etal-2020-learning}):
\begin{itemize}
    \item \textsc{Absolute token position} (ATP): This feature is 1 \textit{iff} \textit{the} is the first token of the sentence.
    \item \textsc{Length} (L): This feature is 1 \textit{iff} the sentence contains more than \textit{n} (3) words.
    \item \textsc{Lexical content} (LCT): This feature is 1 \textit{iff} the sentence contains \textit{the}.
    \item \textsc{Relative token position} (RTP): This feature is 1 when \textit{the} precedes \textit{a}, and 0 when \textit{a} precedes \textit{the}.
    \item \textsc{Orthography} (TC): This feature is 1 \textit{iff} the sentence is in title case.
\end{itemize}
The linguistic features are (definitions taken from \citet{warstadt-etal-2020-learning}):
\begin{itemize}
    \item \textsc{Syntactic construction} (CR): This feature has value 1 \textit{iff} the sentence contains the control construction.
    \item \textsc{Morphology} (IF): This feature is 1 \textit{iff} the sentence contains an irregular verb in the past tense.
    \item \textsc{Syntactic position} (MV): This feature is 1 \textit{iff} the sentence's main verb is in the \textit{-ing} form.
    \item \textsc{Syntactic category} (SC): This feature is 1 \textit{iff} the sentence contains an adjective.
\end{itemize}

\begin{figure}[h!]
    \centering
    \includegraphics[width=\textwidth]{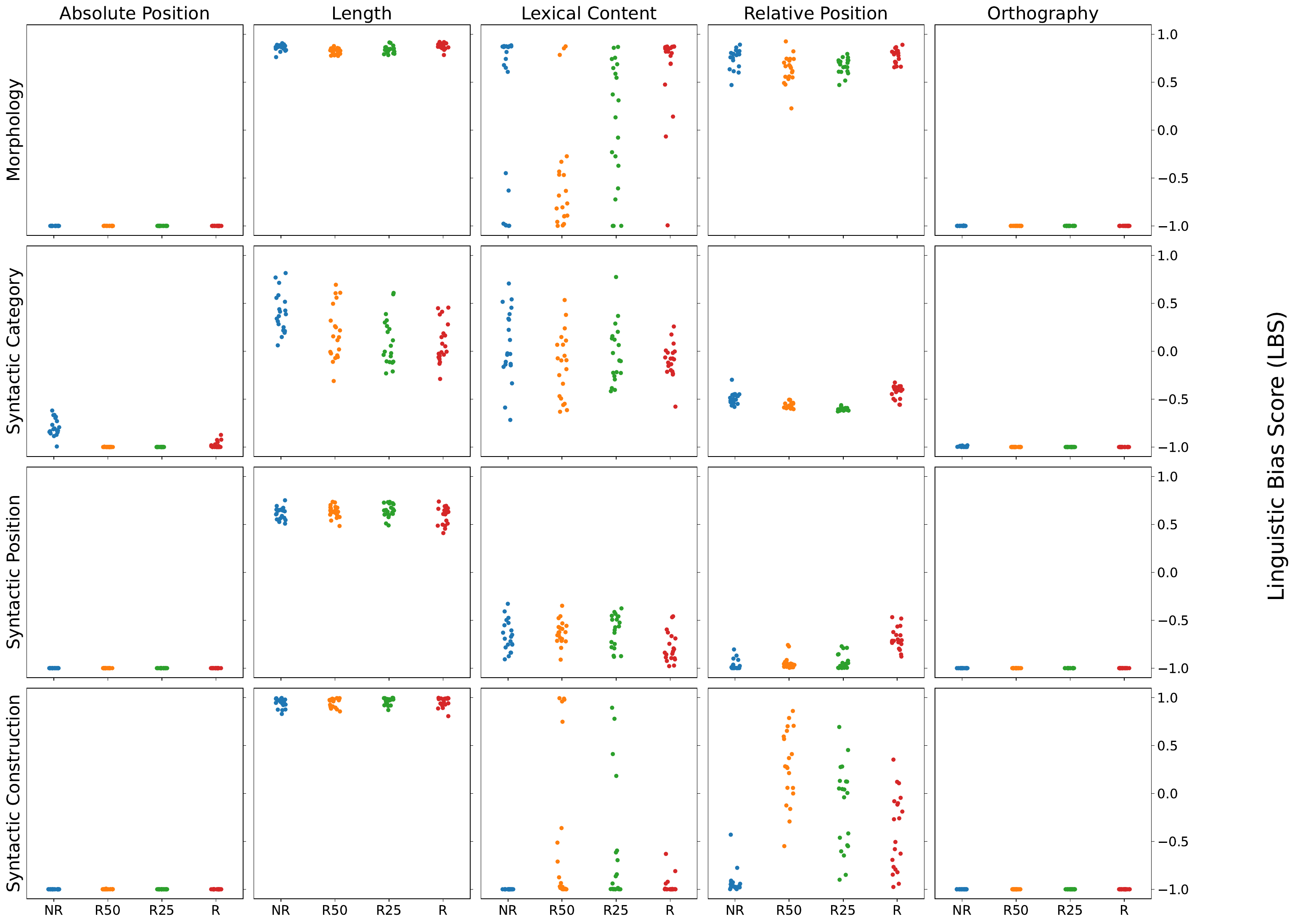}
    \caption{The dots in each sub-plot represent the LBS score of each run of each model. Each model has 20 different runs for each combination of surface and linguistic features. NR = Model pre-trained without retrieval, R50 = Model pre-trained with 50\% noisy retrieval, R25 = Model pre-trained with 25\% noisy retrieval, R = Model pre-trained with 0\% noisy retrieval}
    \label{fig:MSGS}
\end{figure}

For every model, we run five different seeds: 34, 42, 74, 2395, and 10801 at four different learning rates: 1e-5, 3e-5, 5e-5, 1e-4. \cref{fig:MSGS} shows the distribution of all our runs for the base models from \cref{tab:results}. \cref{tab:MSGS} shows the LBS results over each feature. From this table, we see that our retrieval pre-trained models are better at biasing the morphology feature and biasing less the lexical content feature while biasing more the length feature compared to the regular pretrained models. In general, the length task is the hardest surface task to detect while morphology is the easiest linguistic task to detect.

\begin{table*}[!h]
\resizebox{\textwidth}{!}{%
\begin{NiceTabular}{@{}lrrrrr@{\hspace{2em}}rrrr@{\hspace{2em}}r@{}}
\toprule
                    & \Block{1-5}{\textsc{Surface Features}} & & & & &  \Block{1-4}{\textsc{Linguistics Features}}\\
                   \textbf{Model} & \textbf{ATP} & \textbf{L} & \textbf{LCT} & \textbf{RTP} & \textbf{TC} & \textbf{CR} & \textbf{IF} & \textbf{MV} & \textbf{SC} & \textbf{Average} \\ 
\midrule

    \textsc{reference model}& & & \\
 \hspace{1em}\textit{bert-base-cased}& \textit{-0.55} & \textit{0.66} & \textit{0.28} & \textit{0.05} & \textit{-0.95} & \textit{-0.36} & \textit{0.31} & \textit{-0.19} & \textit{-0.17} & \textit{-0.10} \\[0.5em]
 
 \textsc{base}& & & \\
 \hspace{1em}$\bm{-}$ retrieval pretraining (patch) & \underline{-0.96} & \textbf{0.70} & \underline{-0.37} & -0.40 & \textbf{-1.00} & -0.62 & \underline{-0.06} & -0.59 & \textbf{-0.35} & -0.40 \\
 \hspace{1em}$\bm{-}$ retrieval pretraining (no patch) & \textbf{-0.95} & \underline{0.68} & -0.63 & -0.30 & \textbf{-1.00} & -0.62 & -0.20 & -0.57 & \underline{-0.46} & -0.47 \\
 \hspace{1em}$\bm{+}$ retrieval pretraining (50\% noise, patch) & -1.00 & 0.65 & -0.42 & \textbf{-0.07} & \textbf{-1.00} & \textbf{-0.52} & -0.21 & \textbf{-0.24} & -0.50 & $\dagger$ \textbf{-0.37} \\
 \hspace{1em}$\bm{+}$ retrieval pretraining (25\% noise, patch) & -1.00 & 0.64 & \textbf{-0.30} & -0.25 & \textbf{-1.00} & -0.58 & -0.09 & \underline{-0.36} & -0.51 & $\dagger$ \underline{-0.38} \\
 \hspace{1em}$\bm{+}$ retrieval pretraining (0\% noise, patch) & -1.00 & 0.65 & \textbf{-0.30} & \underline{-0.19} & \textbf{-1.00} & -0.58 & \textbf{0.06} & -0.49 & -0.47 & \textbf{-0.37} \\
 \hspace{1em}$\bm{+}$ retrieval pretraining (0\% noise, no patch) & -1.00 & 0.57 & -0.88 & -0.30 & \textbf{-1.00} & \underline{-0.56} & -0.29 & -0.57 & -0.67 & -0.52 \\[0.5em]

 \textsc{small}& & &\\
\hspace{1em}$\bm{-}$ retrieval pretraining (patch) & \textbf{-1.00} & \underline{0.56} & -0.81 & -0.53 & \textbf{-1.00} & \underline{-0.59} & -0.29 & \textbf{-0.62} & -0.73 & -0.56 \\
\hspace{1em}$\bm{-}$ retrieval pretraining (no patch) & \textbf{-1.00} & \textbf{0.59} & -0.77 & \underline{-0.43} & \textbf{-1.00} & \textbf{-0.56} & -0.31 & \textbf{-0.62} & \textbf{-0.60} & \underline{-0.52} \\
\hspace{1em}$\bm{+}$ retrieval pretraining (0\% noise, patch) & \textbf{-1.00} & 0.54 & \underline{-0.75} & \underline{-0.43} & \textbf{-1.00} & -0.60 & \underline{-0.22} & \underline{-0.63} & -0.66 & -0.53 \\
\hspace{1em}$\bm{+}$ retrieval pretraining (0\% noise, no patch) & \textbf{-1.00} & 0.54 & \textbf{-0.66} & \textbf{-0.44} & \textbf{-1.00} & \underline{-0.59} & \textbf{-0.14} & -0.64 & \underline{-0.64} & $\dagger$  \textbf{-0.50} \\[0.5em]

 \textsc{x-small}& & &\\
\hspace{1em}$\bm{-}$ retrieval pretraining (patch) & \textbf{-1.00} & \underline{0.36} & \underline{-0.73} & -0.45 & \textbf{-1.00} & -0.60 & \underline{-0.28} & \underline{-0.67} & \underline{-0.71} & \underline{-0.57} \\
\hspace{1em}$\bm{-}$ retrieval pretraining (no patch) & \textbf{-1.00} & \textbf{0.44} & -0.79 & \textbf{-0.42} & \textbf{-1.00} & -0.60 & -0.30 & \textbf{-0.64} & \textbf{-0.69} & \textbf{-0.55} \\
\hspace{1em}$\bm{+}$ retrieval pretraining (0\% noise, patch) & \textbf{-1.00} & 0.33 & -0.76 & \underline{-0.44} & \textbf{-1.00} & \underline{-0.58} & -0.32 & -0.71 & \textbf{-0.69} & \underline{-0.57} \\ 
\hspace{1em}$\bm{+}$ retrieval pretraining (0\% noise, no patch) & \textbf{-1.00} & 0.22 & \textbf{-0.69} & -0.47 & \textbf{-1.00} & \textbf{-0.56} & \textbf{-0.24} & -0.81 & -0.74 & -0.59 \\

\bottomrule
\end{NiceTabular}%
}
\caption{Fine-grained MSGS results. ATP = Absolute Token Position, L = Length, LCT = Lexical Content, RTP = Relative Token Position, TC = Orthography, CR = Syntactic Construction, IF = Morphology, MV = Syntactic Position, and SC = Syntactic Category. The \textbf{bold} numbers represent the best model at each size, while the \underline{underline} is the second best. $\dagger$ indicates that the result is significantly better than the no-retrieval model based on the ASO test.} 
\label{tab:MSGS}
\end{table*}

\subsection{LAMBADA}
\label{app:lambada}

LAMBADA is a zero-shot language modeling task that focuses on resolving long-range dependencies in text \citep{paperno-etal-2016-lambada}; we used its detokenized version from \newcite{radford2019language}. While it has been traditionally used for evaluating autoregressive language models, we adapt the task for masked language models.\footnote{We made this version of LAMBADA openly available at \url{https://huggingface.co/datasets/ltg/lambada-context}.} Note that this adaptation does not allow for a direct comparison with the autoregressive models. An illustrative sample from this dataset looks as follows:

\textbf{Prompt:} \textit{"Give me a minute to change and I'll meet you at the docks." She'd forced those words through her teeth. "No need to change. We won't be that long." Shane gripped her arm and started leading her to the dock. "I can make it there on my own, \textbf{\{answer\}}."}

\textbf{Gold answer:} \textit{Shane}

We insert the whole tokenized prompt to the evaluated language model and replace the missing answer by $k$ mask tokens, where $k$ is the length of the tokenized gold answer. Then we evaluate the exact-match accuracy of predicting filling in the correct continuation and also the mean perplexity.

\begin{table*}[!h]
\small
\centering
\begin{tabular}{@{}lrr@{}}
\toprule
\textbf{Model} & \textbf{Accuracy} & \textbf{Perplexity} \\ 
\midrule
    \textsc{reference model} \\
 \hspace{1em}\textit{bert-base-cased} & \textit{44.77} & \textit{26.95} \\[0.5em]
 \textsc{base} \\
 \hspace{1em}$\bm{-}$ retrieval pretraining (patch) & \textbf{47.00} & \textbf{17.60} \\
\hspace{1em}$\bm{-}$ retrieval pretraining (no patch) & \underline{46.09} & \underline{18.56} \\
\hspace{1em}$\bm{+}$ retrieval pretraining (50\% noise, patch) & 43.22 & 24.40 \\
\hspace{1em}$\bm{+}$ retrieval pretraining (25\% noise, patch) & 40.58 & 29.62 \\
\hspace{1em}$\bm{+}$ retrieval pretraining (0\% noise, patch) & 37.59 & 39.84 \\
  \hspace{1em}$\bm{+}$ retrieval pretraining (0\% noise, no patch) & 22.63 & 141.62  \\ [0.5em]
 \textsc{small} \\
\hspace{1em}$\bm{-}$ retrieval pretraining (patch) & 
\underline{35.11} & \underline{44.81} \\
\hspace{1em}$\bm{-}$ retrieval pretraining (no patch) & \textbf{35.84} & \textbf{41.25} \\
\hspace{1em}$\bm{+}$ retrieval pretraining (0\% noise, patch) & 26.24 & 135.94 \\
\hspace{1em}$\bm{+}$ retrieval pretraining (0\% noise, no patch) & 0.43 & 37183.08 \\ [0.5em]
 \textsc{x-small} \\
\hspace{1em}$\bm{-}$ retrieval pretraining (patch) & \textbf{25.42} & \textbf{133.44} \\
\hspace{1em}$\bm{-}$ retrieval pretraining (no patch) & \underline{25.33} & \underline{137.73} \\
\hspace{1em}$\bm{+}$ retrieval pretraining (0\% noise, patch) & 19.33 & 329.90 \\
\hspace{1em}$\bm{+}$ retrieval pretraining (0\% noise, no patch) & 0.00 & $1.88\times10^{11}$\\

\bottomrule
\end{tabular}
\caption{Fine-grained LAMBADA results. The \textbf{bold} numbers represent the best model in each size, while the \underline{underline} is the second best.}
\label{tab:lambada}
\end{table*}


\subsection{GLUE}
\label{app:glue}

To judge one of the facets of language understanding we use most of the GLUE benchmark \citep{wang2019glue}. The benchmark is composed of the following tasks:

\begin{itemize}\itemsep0em 
    \item \textbf{Corpus of Linguistic Acceptability} \citep[CoLA;][]{warstadt-etal-2019-neural} evaluated with the Matthews correlation coefficient \citep[MCC;][]{MATTHEWS1975442}.
    \item \textbf{The Stanford Sentiment Treebank} \citep[SST-2;][]{socher-etal-2013-recursive}, evaluated with accuracy.
    \item \textbf{The Microsoft Research Paraphrase Corpus} \citep[MRPC;][]{dolan-brockett-2005-automatically}, evaluated with both F\textsubscript{1}-score (originally also evaluated with accuracy).
    \item \textbf{The Quora Question Pairs} (QQP),\footnote{\url{https://quoradata.quora.com/First-Quora-Dataset-Release-Question-Pairs}} evaluated with F\textsubscript{1}-score (originally evaluated with accuracy).
    \item \textbf{The Multi-Genre Natural Language Inference Corpus} \citep[MNLI;][]{williams-etal-2018-broad}. Its development set consists of two parts: \textit{matched}, sampled from the same data source as the training set, and \textit{mismatched}, which is sampled from a different domain. Both parts are evaluated with accuracy.
    \item \textbf{Question-answering Natural Language Inference} (QNLI) constructed from the Stanford Question Answering Dataset \citep[SQuAD;][]{rajpurkar-etal-2016-squad}, evaluated with accuracy.
    \item \textbf{The Recognizing Textual Entailment datasets} \citep[RTE;][]{10.1007/11736790_9, rte2, giampiccolo-etal-2007-third, Bentivogli09thefifth}, evaluated with accuracy.
    \item \textbf{The Semantic Textual Similarity Benchmark} \citep[STS-B;][]{cer-etal-2017-semeval} is a collection of sentence pairs drawn from news headlines, video and image captions, and natural language inference data. Each pair is human-annotated with a similarity score from 1 to 5; the task is to predict these scores. We evaluate using Pearson and Spearman correlation coefficients.
    \item \textbf{Winograd Schema Challenge} \citep[WSC;][]{levesque2011winograd} evaluated with accuracy.
\end{itemize}

We omit the Winograd Schema Challenge due to the lack of training and test data leading to all our models underperforming compared to the majority label.

\cref{tab:glue} shows the detailed results of each of the GLUE tasks. We see that independent of model size, the retrieval pre-trained models perform better on the CoLA dataset, although the difference between the models shrinks as the model size grows. In addition, we see inversions in the MNLI, RTE and STS-B tasks with the XS model performing better, the Small model on par and the Base model performing worse.

We did an extensive hyperparameter search for the retrieval pre-trained patched base and xs models as well as the regular pre-trained base and xs models. For the small version, we limited our learning rates to be in between those of the base and xs models. For the noisy versions, we combined the hyperparameters of the retrieval and regular pre-trained model and divided them by the amount of noise. In other words, the values of the learning rate for 25\% noise are 25\% of the way from the retrieval parameters going to the regular parameters, while keeping the batch size and warmup ratio the same as the retrieval version (although we made a mistake and did the opposite but to save compute, we have not re-run them correctly). For the 50\% noise, we took the half-point values for all three hyperparameters. Finally, we used the hyperparameters of the base regular pre-trained models for \textsc{bert-base-cased}. The detailed list of the hyperparameters can be found in \cref{tab:glue_hyp}.

\vspace{1em}

\begin{table*}[!h]
\resizebox{\textwidth}{!}{%
\begin{tabular}{@{}lrrrrrrrrrr@{}}
\toprule
                   \textbf{Model} & \textbf{CoLA} & \textbf{SST-2} & \textbf{MRPC} & \textbf{QQP} & \textbf{MNLI} & \textbf{MNLI-mm} & \textbf{QNLI} & \textbf{RTE} & \textbf{STS-B} & \textbf{Average} \\ 
\midrule

    \textsc{reference model}& & & \\
 \hspace{1em}\textit{bert-base-cased}& \textit{57.4}$^{\pm0.6}$ & \textit{91.3}$^{\pm0.5}$ & \textit{89.2}$^{\pm0.6}$ & \textit{87.2}$^{\pm0.2}$ & \textit{82.5}$^{\pm0.3}$ & \textit{82.9}$^{\pm0.3}$ & \textit{89.2}$^{\pm0.2}$ & \textit{63.9}$^{\pm3.5}$ & \textit{88.9}$^{\pm0.6}$/\textit{88.5}$^{\pm0.7}$ & \textit{82.1}$^{\pm1.2}$ \\[0.5em]
 
 \textsc{base}& & & \\
 \hspace{1em}$\bm{-}$ retrieval pretraining (patch) & \underline{51.9}$^{\pm1.1}$ & \textbf{91.8}$^{\pm0.9}$ & \underline{90.5}$^{\pm0.4}$ & \textbf{88.2}$^{\pm0.1}$ & \underline{84.2}$^{\pm0.2}$ & \textbf{84.4}$^{\pm0.3}$ & \underline{91.4}$^{\pm0.3}$ & 62.1$^{\pm3.8}$ & \textbf{87.9}$^{\pm0.3}$/\textbf{87.7}$^{\pm0.3}$ & \underline{82.0}$^{\pm1.3}$ \\
\hspace{1em}$\bm{-}$ retrieval pretraining (no patch) & \underline{51.9}$^{\pm1.5}$ & 91.3$^{\pm0.5}$ & \textbf{90.6}$^{\pm0.5}$ & \textbf{88.2}$^{\pm0.2}$ & \textbf{84.4}$^{\pm0.1}$ & \textbf{84.4}$^{\pm0.2}$ & \textbf{91.5}$^{\pm0.2}$ & \underline{64.4}$^{\pm3.9}$ & \underline{87.8}$^{\pm0.4}$/\underline{87.6}$^{\pm0.4}$ & \textbf{82.2}$^{\pm1.4}$ \\
\hspace{1em}$\bm{+}$ retrieval pretraining (50\% noise, patch) & 51.7$^{\pm1.5}$ & 91.2$^{\pm0.9}$ & 90.3$^{\pm0.9}$ & \underline{88.0}$^{\pm0.1}$ & 83.9$^{\pm0.1}$ & \underline{83.9}$^{\pm0.1}$ & 91.3$^{\pm0.1}$ & \textbf{64.9}$^{\pm3.5}$ & 87.7$^{\pm0.3}$/87.5$^{\pm0.3}$ & \underline{82.0}$^{\pm1.3}$ \\
\hspace{1em}$\bm{+}$ retrieval pretraining (25\% noise, patch) & 51.8$^{\pm0.5}$ & \underline{91.4}$^{\pm0.2}$ & \textbf{90.6}$^{\pm0.6}$ & 87.9$^{\pm0.1}$ & 83.9$^{\pm0.3}$ & 83.8$^{\pm0.2}$ & 91.1$^{\pm0.1}$ & 63.5$^{\pm1.4}$ & 87.7$^{\pm0.4}$/87.4$^{\pm0.4}$ & 81.9$^{\pm0.6}$ \\
\hspace{1em}$\bm{+}$ retrieval pretraining (0\% noise, patch) & 51.4$^{\pm1.8}$ & 91.3$^{\pm0.8}$ & 90.1$^{\pm1.2}$ & 87.8$^{\pm0.2}$ & 83.3$^{\pm0.1}$ & 83.4$^{\pm0.2}$ & 90.2$^{\pm0.3}$ & 61.1$^{\pm3.6}$ & 86.8$^{\pm0.3}$/86.6$^{\pm0.3}$ & 81.2$^{\pm1.4}$ \\
\hspace{1em}$\bm{+}$ retrieval pretraining (0\% noise, no patch) & \textbf{53.1}$^{\pm0.4}$ & 90.6$^{\pm0.4}$ & 88.0$^{\pm1.0}$ & 87.8$^{\pm0.1}$ & 83.2$^{\pm0.2}$ & 83.4$^{\pm0.3}$ & 89.5$^{\pm0.2}$ & 55.8$^{\pm1.7}$ & 86.5$^{\pm0.3}$/86.1$^{\pm0.3}$ & 80.4$^{\pm0.7}$ \\ [0.5em]

 \textsc{small}& & &\\
\hspace{1em}$\bm{-}$ retrieval pretraining (patch) & 35.3$^{\pm1.8}$ & 89.1$^{\pm0.8}$ & \underline{88.3}$^{\pm1.2}$ & \underline{86.6}$^{\pm0.1}$ & 81.7$^{\pm0.2}$ & \underline{82.0}$^{\pm0.3}$ & \underline{89.4}$^{\pm0.5}$ & \underline{53.4}$^{\pm3.3}$ & 84.2$^{\pm0.5}$/83.8$^{\pm0.5}$ & 77.4$^{\pm1.3}$ \\
\hspace{1em}$\bm{-}$ retrieval pretraining (no patch) & 37.5$^{\pm2.8}$ & \underline{89.8}$^{\pm0.5}$ & \textbf{88.4}$^{\pm0.7}$ & \textbf{86.9}$^{\pm0.1}$ & \textbf{82.0}$^{\pm0.1}$ & \textbf{82.6}$^{\pm0.1}$ & \textbf{89.5}$^{\pm0.3}$ & 53.3$^{\pm2.3}$ & \textbf{85.1}$^{\pm0.5}$/\textbf{84.7}$^{\pm0.5}$ & \underline{78.0}$^{\pm1.2}$ \\
\hspace{1em}$\bm{+}$ retrieval pretraining (0\% noise, patch) & \underline{40.4}$^{\pm2.1}$ & \textbf{90.6}$^{\pm0.5}$ & \underline{88.3}$^{\pm1.2}$ & \underline{86.6}$^{\pm0.1}$ & \underline{81.8}$^{\pm0.2}$ & \underline{82.0}$^{\pm0.2}$ & 89.0$^{\pm0.3}$ & \textbf{55.8}$^{\pm1.4}$ & \textbf{85.1}$^{\pm0.4}$/\textbf{84.7}$^{\pm0.4}$ & $\dagger$ \textbf{78.5}$^{\pm0.9}$ \\
\hspace{1em}$\bm{+}$ retrieval pretraining (0\% noise, no patch) & \textbf{40.9}$^{\pm1.8}$ & 89.7$^{\pm0.4}$ & 86.5$^{\pm0.6}$ & 86.5$^{\pm0.2}$ & 81.5$^{\pm0.3}$ & 81.9$^{\pm0.3}$ & 87.8$^{\pm0.4}$ & \underline{53.4}$^{\pm2.0}$ & \underline{84.4}$^{\pm0.5}$/\underline{84.1}$^{\pm0.4}$ & 77.7$^{\pm0.9}$ \\ [0.5em]

 \textsc{x-small}& & &\\
\hspace{1em}$\bm{-}$ retrieval pretraining (patch) &  \underline{25.5}$^{\pm1.5}$ & 88.1$^{\pm0.5}$ & \underline{88.3}$^{\pm0.7}$ & 84.6$^{\pm0.2}$ & 78.3$^{\pm0.2}$ & 79.3$^{\pm0.2}$ & 86.4$^{\pm0.2}$ & 51.1$^{\pm4.7}$ & 82.4$^{\pm0.5}$/82.0$^{\pm0.5}$ & 74.6$^{\pm1.6}$ \\
\hspace{1em}$\bm{-}$ retrieval pretraining (no patch) & 25.0$^{\pm3.7}$ & \underline{88.6}$^{\pm0.4}$ & \textbf{88.7}$^{\pm0.9}$ & \textbf{85.0}$^{\pm0.1}$ & 78.8$^{\pm0.3}$ & 79.7$^{\pm0.1}$ & \textbf{86.9}$^{\pm0.4}$ & \underline{54.1}$^{\pm1.4}$ & \underline{82.8}$^{\pm0.2}$/\underline{82.3}$^{\pm0.2}$ & \underline{75.2}$^{\pm1.3}$ \\
\hspace{1em}$\bm{+}$ retrieval pretraining (0\% noise, patch) & \textbf{32.7}$^{\pm2.4}$ & \underline{88.6}$^{\pm0.7}$ & 87.3$^{\pm1.0}$ & \underline{84.9}$^{\pm0.1}$ & \textbf{79.6}$^{\pm0.3}$ & \underline{80.0}$^{\pm0.3}$ & \underline{86.8}$^{\pm0.2}$ & \textbf{55.4}$^{\pm2.2}$ & 82.5$^{\pm0.7}$/\underline{82.3}$^{\pm0.7}$ & $\dagger$  \textbf{76.0}$^{\pm1.1}$ \\
\hspace{1em}$\bm{+}$ retrieval pretraining (0\% noise, no patch) & 25.4$^{\pm2.2}$ & \textbf{89.0}$^{\pm0.6}$ & 85.0$^{\pm1.0}$ & 84.7$^{\pm0.2}$ & \underline{79.5}$^{\pm0.1}$ & \textbf{80.2}$^{\pm0.2}$ & 85.2$^{\pm0.5}$ & 52.0$^{\pm3.3}$ & \textbf{82.9}$^{\pm0.4}$/\textbf{82.7}$^{\pm0.4}$ & 74.6$^{\pm1.3}$ \\

\bottomrule
\end{tabular}%
}
\caption{Fine-grained GLUE results. The CoLA metric is MCC, the F1-score is used for MRPC and QQP, and the other tasks are evaluated with accuracy. The results are reported as the mean and the standard deviation from 5 seeded runs. The \textbf{bold} numbers represent the best model at each size, while the \underline{underline} is the second best. $\dagger$ indicates that the result is significantly better than the no-retrieval model based on the ASO test.} 
\label{tab:glue}
\end{table*}

\begin{table*}[!h]
\resizebox{\textwidth}{!}{%
\begin{tabular}{@{}lrrrrrrrr@{}}
\toprule
                   \textbf{Hyperparameter} & \textbf{CoLA} & \textbf{SST-2} & \textbf{MRPC} & \textbf{QQP} & \textbf{MNLI} & \textbf{QNLI} & \textbf{RTE} & \textbf{STS-B}\\ 
\midrule
\textsc{shared} & & & \\
 \hspace{1em}Epochs & 10 & 10 & 10 & 4 & 4 & 10 & 10 & 10 \\
 \hspace{1em}Weight decay & 0.1 & 0.1 & 0.1 & 0.1 & 0.1 & 0.1 & 0.1 & 0.1 \\
 \hspace{1em}Learning Rate Scheduler & linear & linear & linear & linear & linear & linear & linear & linear \\
 \hspace{1em}Attention Dropout & 0.1 & 0.1 & 0.1 & 0.1 & 0.1 & 0.1 & 0.1 & 0.1 \\
 \hspace{1em}Classifier Dropout & 0.1 & 0.1 & 0.1 & 0.1 & 0.1 & 0.1 & 0.1 & 0.1 \\
 \hspace{1em}Adam Epsilon & 1e-6 & 1e-6 & 1e-6 & 1e-6 & 1e-6 & 1e-6 & 1e-6 & 1e-6 \\ [0.5em]
\textsc{base $\bm{-}$ retrieval \& reference model}& & & \\
 \hspace{1em}Learning rate & 2e-5 & 2e-5 & 5e-5 & 5e-5 & 5e-5 & 5e-5 & 1e-4 & 1.2e-4 \\
 \hspace{1em}Batch size & 16 & 16 & 16 & 16 & 16 & 16 & 32 & 32 \\
 \hspace{1em}Warmup Ratio & 0.1 & 0.06 & 0.1 & 0.06 & 0.1 & 0.06 & 0.06 & 0.1 \\ [0.5em]
\textsc{base $\bm{+}$ retrieval(50\% noise)}& & & \\
 \hspace{1em}Learning rate & 3.5e-5 & 2e-5 & 7.5e-5 & 5e-5 & 5e-5 & 3.5e-5 & 1e-4 & 1.35e-4 \\
 \hspace{1em}Batch size & 24 & 16 & 24 & 16 & 24 & 16 & 32 & 24 \\
 \hspace{1em}Warmup Ratio & 0.08 & 0.06 & 0.1 & 0.08 & 0.1 & 0.08 & 0.06 & 0.1 \\ [0.5em]
\textsc{base $\bm{+}$ retrieval (25\% noise)}& & & \\
 \hspace{1em}Learning rate & 2.75e-5 & 2e-5 & 6.25e-5 & 5e-5 & 5e-5 & 4.25e-5 & 1e-4 & 1.275e-4 \\
 \hspace{1em}Batch size & 16 & 16 & 16 & 16 & 16 & 16 & 32 & 32 \\
 \hspace{1em}Warmup Ratio & 0.1 & 0.06 & 0.1 & 0.06 & 0.1 & 0.06 & 0.06 & 0.1 \\ [0.5em]
\textsc{base $\bm{+}$ retrieval}& & & \\
 \hspace{1em}Learning rate & 5e-5 & 2e-5 & 1e-4 & 5e-5 & 5e-5 & 2e-5 & 1e-4 & 1.5e-4 \\
 \hspace{1em}Batch size & 32 & 16 & 32 & 16 & 32 & 16 & 32 & 16 \\
 \hspace{1em}Warmup Ratio & 0.06 & 0.06 & 0.1 & 0.1 & 0.1 & 0.1 & 0.06 & 0.1 \\ [0.5em]
\textsc{small $\bm{-}$ retrieval}& & & \\
 \hspace{1em}Learning rate & 1.5e-4 & 2e-4 & 1e-4 & 1.5e-4 & 1e-4 & 5e-5 & 1e-4 & 1.8e-4 \\
 \hspace{1em}Batch size & 32 & 32 & 8 & 32 & 32 & 16 & 8 & 8 \\
 \hspace{1em}Warmup Ratio & 0.03 & 0.1 & 0.1 & 0.06 & 0.1 & 0.06 & 0.03 & 0.06 \\ [0.5em]
\textsc{small $\bm{+}$ retrieval}& & & \\
 \hspace{1em}Learning rate & 1e-4 & 1e-4 & 1.25e-4 & 1e-4 & 1e-4 & 3e-5 & 1.25e-4 & 2e-4 \\
 \hspace{1em}Batch size & 32 & 32 & 16 & 16 & 32 & 16 & 16 & 32 \\
 \hspace{1em}Warmup Ratio & 0.03 & 0.06 & 0.06 & 0.06 & 0.06 & 0.06 & 0.06 & 0.12 \\ [0.5em]
\textsc{xs $\bm{-}$ retrieval}& & & \\
 \hspace{1em}Learning rate & 1.5e-4 & 2e-4 & 1e-4 & 1.5e-4 & 2e-4 & 5e-5 & 5e-5 & 2e-4 \\
 \hspace{1em}Batch size & 16 & 16 & 32 & 16 & 32 & 16 & 8 & 8 \\
 \hspace{1em}Warmup Ratio & 0.1 & 0.1 & 0.06 & 0.1 & 0.15 & 0.06 & 0.06 & 0.03 \\ [0.5em]
\textsc{xs $\bm{+}$ retrieval}& & & \\
 \hspace{1em}Learning rate & 1e-4 & 2.8e-4 & 1.5e-4 & 2.2e-4 & 1.8e-4 & 5e-5 & 1.5e-4 & 2e-4 \\
 \hspace{1em}Batch size & 8 & 32 & 16 & 32 & 32 & 16 & 16 & 32 \\
 \hspace{1em}Warmup Ratio & 0.12 & 0.1 & 0.06 & 0.06 & 0.1 & 0.1 & 0.06 & 0.06\\

\bottomrule
\end{tabular}%
}
\caption{Fine-tuning hyperparameter details of GLUE, these are the optimal values found by the grid search described in \cref{app:glue}.} 
\label{tab:glue_hyp}
\end{table*}


\subsection{SQuAD}
\label{app:squad}
SQuAD is an extractive question answering dataset with 107,785 question-answer pairs. The task is to answer questions by providing the span of the correct answer string from a provided passage that is known to answer the question. We finetune all models over three epochs, using a learning rate of $5e-5$, a batch size of $16$, and a weight decay of $0.01$. Models are evaluated on the original development set, with no additional data used. We report the percentage of token-level exact matches (EM) and F1-score. The full set of results can be seen in \cref{tab:squad}.

We observe that retrieval impairs performance for all model sizes. For the base versions, the absolute performance decrease follow the amount of retrieved documents given to the model, showing that the closer one gets to a "perfect" set of retrieved documents, the worse the language model performs on the task of extractive QA. Furthermore, we observe that the addition of our patched linear layer has little effect on SQuAD for all model sizes, which we hypothesize is due to the size of the dataset; with over 100k examples, finetuning allows the model to fully "recover", making the patch obsolete. 

\vspace{1em}

\begin{table*}[!h]
\small
\centering
\begin{tabular}{@{}lrr@{}}
\toprule
                   \textbf{Model} & \textbf{Exact Match} & \textbf{F\textsubscript1 score}  \\ 
\midrule

    \textsc{reference model} \\
 \hspace{1em}\textit{bert-base-cased}& \textit{80.6}$^{\pm0.2}$ & \textit{88.4}$^{\pm0.3}$ \\[0.5em]
 
 \textsc{base} \\
 \hspace{1em}$\bm{-}$ retrieval pretraining (patch) &
 \textbf{84.6}$^{\pm0.2}$ & \textbf{91.3}$^{\pm0.1}$ \\
\hspace{1em}$\bm{-}$ retrieval pretraining (no patch) & $\dagger$ \underline{84.4}$^{\pm0.4}$ & $\dagger$ \underline{91.2}$^{\pm0.2}$ \\
\hspace{1em}$\bm{+}$ retrieval pretraining (50\% noise, patch) & 83.9$^{\pm0.1}$ & 90.7$^{\pm0.2}$  \\
\hspace{1em}$\bm{+}$ retrieval pretraining (25\% noise, patch) & 83.3$^{\pm0.5}$ & 90.2$^{\pm0.2}$ \\
\hspace{1em}$\bm{+}$ retrieval pretraining (0\% noise, patch) & 82.8$^{\pm0.1}$ & 89.7$^{\pm0.2}$ \\
\hspace{1em}$\bm{+}$ retrieval pretraining (0\% noise, no patch) & 82.2$^{\pm0.1}$ & 89.7$^{\pm0.2}$ \\ [0.5em]

 \textsc{small} \\
\hspace{1em}$\bm{-}$ retrieval pretraining (patch) & \underline{81.5}$^{\pm0.2}$ & \textbf{88.6}$^{\pm0.2}$ \\
\hspace{1em}$\bm{-}$ retrieval pretraining (no patch) & $\dagger$ \textbf{81.7}$^{\pm0.3}$ & $\dagger$ \textbf{88.6}$^{\pm0.2}$ \\
\hspace{1em}$\bm{+}$ retrieval pretraining (0\% noise, patch) & 78.9$^{\pm0.1}$ & \underline{86.3}$^{\pm0.2}$ \\
\hspace{1em}$\bm{+}$ retrieval pretraining (0\% noise, no patch) & 78.9$^{\pm0.1}$ & 86.2$^{\pm0.2}$ \\ [0.5em]

 \textsc{x-small} \\
\hspace{1em}$\bm{-}$ retrieval pretraining (patch) &  \underline{73.5}$^{\pm0.2}$ & \textbf{81.8}$^{\pm0.2}$ \\
\hspace{1em}$\bm{-}$ retrieval pretraining (no patch) & $\dagger$ \textbf{73.6}$^{\pm0.3}$ & $\dagger$ \textbf{81.8}$^{\pm0.2}$ \\
\hspace{1em}$\bm{+}$ retrieval pretraining (0\% noise, patch) & 69.9$^{\pm0.2}$ & \underline{78.7}$^{\pm0.1}$ \\
\hspace{1em}$\bm{+}$ retrieval pretraining (0\% noise, no patch) & 70.0$^{\pm0.2}$ & \underline{78.7}$^{\pm0.1}$ \\

\bottomrule
\end{tabular}%
\caption{Results on SQuAD 1.1. Results are reported as the mean and standard deviation over three random seeds. The \textbf{bold} numbers represent the best model at each size, while the \underline{underline} is the second best. $\dagger$ indicates that the result is significantly better than the retrieval model (no noise, patch) based on the ASO test.} 
\label{tab:squad}
\end{table*}

\section{Retrieval effect - Long-range context resolution}
\label{app:long_context}

Using the LAMBADA task, we evaluate whether using the paraphrase encoder as retrieval helps the model understand long-range context dependencies. To this end, we use the LAMBADA prompt without the answer as "paraphrase" to encode and pass through cross-attention to the encoder model. The results can be seen in \cref{tab:long_context}.

\begin{table*}[!h]
\small
\centering
\begin{tabular}{@{}lrr@{}}
\toprule
\textbf{Model} & \textbf{Accuracy} & \textbf{Perplexity} \\ 
\midrule
 \textsc{base} \\
\hspace{1em}$\bm{-}$ retrieval pretraining & \textbf{46.09} & \textbf{18.56} \\
\hspace{1em}$\bm{+}$ retrieval pretraining (retrieval-augmented) & \underline{40.91} & \underline{30.46} \\
\hspace{1em}$\bm{+}$ retrieval pretraining (patch) & 37.59 & 39.84 \\
 \textsc{small} \\
\hspace{1em}$\bm{-}$ retrieval pretraining & \textbf{35.84} & \textbf{41.25} \\
\hspace{1em}$\bm{+}$ retrieval pretraining (retrieval-augmented) & \underline{32.49} & \underline{71.85} \\
\hspace{1em}$\bm{+}$ retrieval pretraining (patch) & 26.24 & 135.94 \\ [0.5em]
 \textsc{x-small} \\
\hspace{1em}$\bm{-}$ retrieval pretraining & \underline{25.33} & \textbf{137.73} \\
\hspace{1em}$\bm{+}$ retrieval pretraining (retrieval-augmented) & \textbf{29.26} & \underline{160.45} \\
\hspace{1em}$\bm{+}$ retrieval pretraining (patch) & 19.33 & 329.90\\

\bottomrule
\end{tabular}
\caption{Fine-grained LAMBADA results of the patched and retrieval-augmented retrieval pre-trained models and the model trained without retrieval pre-training. We used the prompt without the answer as the retrieved text for the retrieval-augmented models. The \textbf{bold} numbers represent the best model in each size, while the \underline{underline} is the second best.}
\label{tab:long_context}
\end{table*}

As we can see, the results show that the retrieval component of the model potentially encodes long-range context dependencies. We also ran the retrieval-augmented models with the full prompt as retrieved text and got performances close to 1 perplexity (1.11 -- 1.04) and accuracy near 100\% (97.15 -- 98.74).

\end{document}